\documentclass{article}

\PassOptionsToPackage{numbers, compress}{natbib}
\PassOptionsToPackage{table}{xcolor}

\usepackage[pagebackref,breaklinks,colorlinks]{hyperref}

    \usepackage[preprint]{neurips_2023}
    \usepackage{pdfpages}

\usepackage[utf8]{inputenc} 
\usepackage[T1]{fontenc}    
\usepackage{hyperref}       
\usepackage{url}            
\usepackage{booktabs}       
\usepackage{amsfonts}       
\usepackage{nicefrac}       
\usepackage{microtype}      
\usepackage{xcolor}         
\usepackage{tabularray}
\usepackage{caption}
\usepackage{soul}
\usepackage{graphicx}
\usepackage{float}
\usepackage{changepage}
\usepackage{adjustbox}
\usepackage{microtype}
\usepackage{diagbox}
\usepackage[table]{xcolor}

\usepackage{multirow}
\usepackage{multicol}
\usepackage{subcaption}
\usepackage{kotex}
\usepackage{tikz}

\title{TWLV-I: Analysis and Insights from Holistic Evaluation on Video Foundation Models}

\author{
\Large \adjustbox{valign=t}{\includegraphics[width=0.03\linewidth]{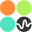}} Twelve Labs
}

\begin{document}

\maketitle

\newcommand{\ourmodel}{TWLV-I}
\definecolor{tl}{RGB}{154,237,89}

\begin{abstract}
In this work, we discuss evaluating video foundation models in a fair and robust manner. Unlike language or image foundation models, many video foundation models are evaluated with differing parameters (such as sampling rate, number of frames, pretraining steps, etc.), making fair and robust comparisons challenging. Therefore, we present a carefully designed evaluation framework for measuring two core capabilities of video comprehension: appearance and motion understanding. Our findings reveal that existing video foundation models, whether text-supervised like UMT or InternVideo2, or self-supervised like V-JEPA, exhibit limitations in at least one of these capabilities. As an alternative, we introduce \textbf{{\ourmodel}}, a new video foundation model that constructs robust visual representations for both motion- and appearance-based videos\footnote{Please cite this paper as (Twelve Labs, 2024). Please see the Authorship Section at the end of this report for the full list of contributors.}. Based on the average top-1 accuracy of linear probing on five action recognition benchmarks, pretrained only on publicly accessible datasets\footnote{This work is strictly intended for academic purposes. Note that Twelve Labs' Embedding API is powered by a model that excludes all non-commercial data.}, our model shows a 4.6\%p improvement compared to V-JEPA~(ViT-L) and a 7.7\%p improvement compared to UMT~(ViT-L). Even when compared to much larger models, our model demonstrates a 7.2\%p improvement compared to DFN~(ViT-H), a 2.7\%p improvement compared to V-JEPA~(ViT-H) and a 2.8\%p improvement compared to InternVideo2~(ViT-g). We provide embedding vectors obtained by {\ourmodel} from videos of several commonly used video benchmarks, along with evaluation source code that can directly utilize these embeddings. The code is available on \url{https://github.com/twelvelabs-io/video-embeddings-evaluation-framework}.
\end{abstract}

\begin{figure}[h]
    \centering
    \begin{subfigure}[b]{0.49\textwidth}
        \centering
        \includegraphics[width=\textwidth]{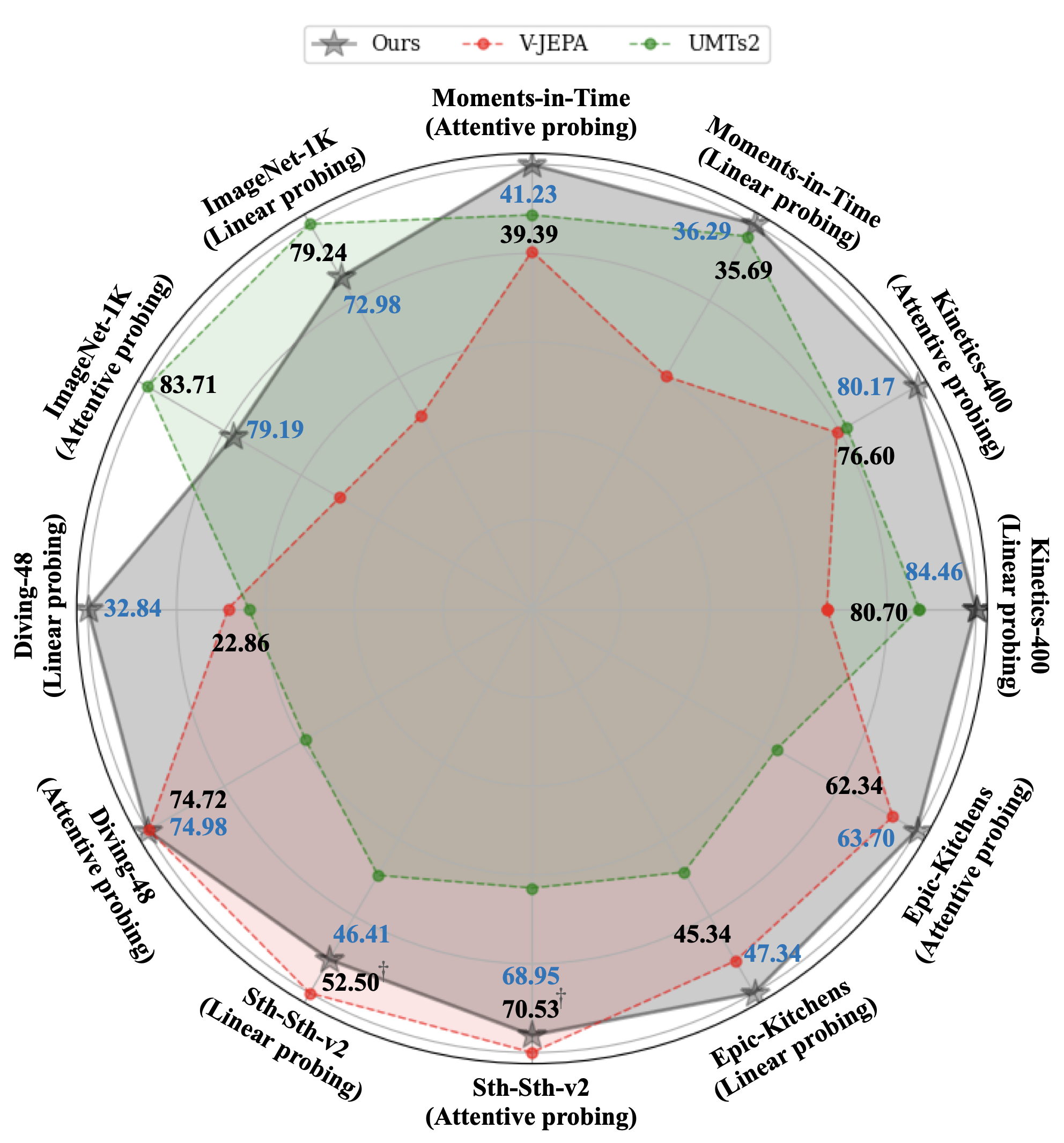}
    \end{subfigure}
    \begin{subfigure}[b]{0.49\textwidth}
        \centering
        \includegraphics[width=\textwidth]{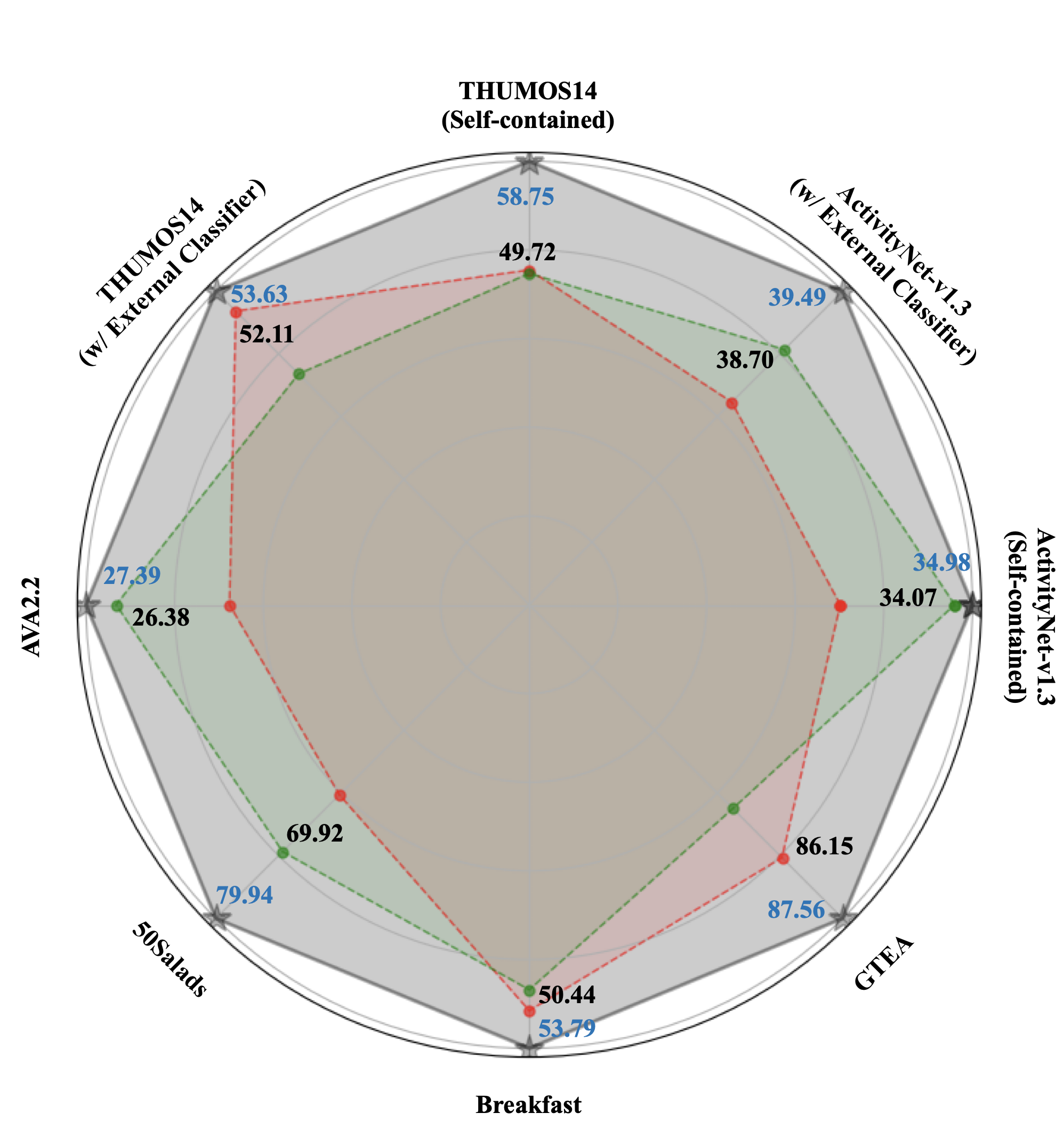}
    \end{subfigure}
    \caption{\textbf{Comparison with video foundation models in the same scale.} \ourmodel~can transfer to various tasks and perform comparable or even superior to the state-of-the-art models. We present the best performance among the competitors, as well as the performance of our model. \textsuperscript{\textdagger} denotes the model that uses the dataset in the pretraining stage. All compared models in this figure are on the ViT-L scale.}
    \label{fig:main_plot}
\end{figure}
\section{Introduction}
\label{sec:intro}

Video is everyone’s first language. From the moment humans are born, they learn about the world by seeing videos even before using language.
Therefore, similar to human languages, developing a video understanding system is essential to achieve the ability to perceive the world accurately. Since videos are sequences of images, it is crucial to recognize what appears in each frame (\textbf{appearance}). In addition to appearance, videos contain intrinsic characteristics not present in images: \textbf{motion}. In this technical report, we consolidate and improve existing evaluation methods for video understanding, while also introducing several new methodologies to propose a comprehensive and fair evaluation framework from the perspectives of appearance and motion. Along with the evaluation framework, we introduce a video foundation model that can comprehensively understand both the appearance and motion in videos.

\begin{figure}[t]
    \centering
    \begin{subfigure}[b]{0.47\textwidth}
        \centering
        \includegraphics[width=\textwidth]{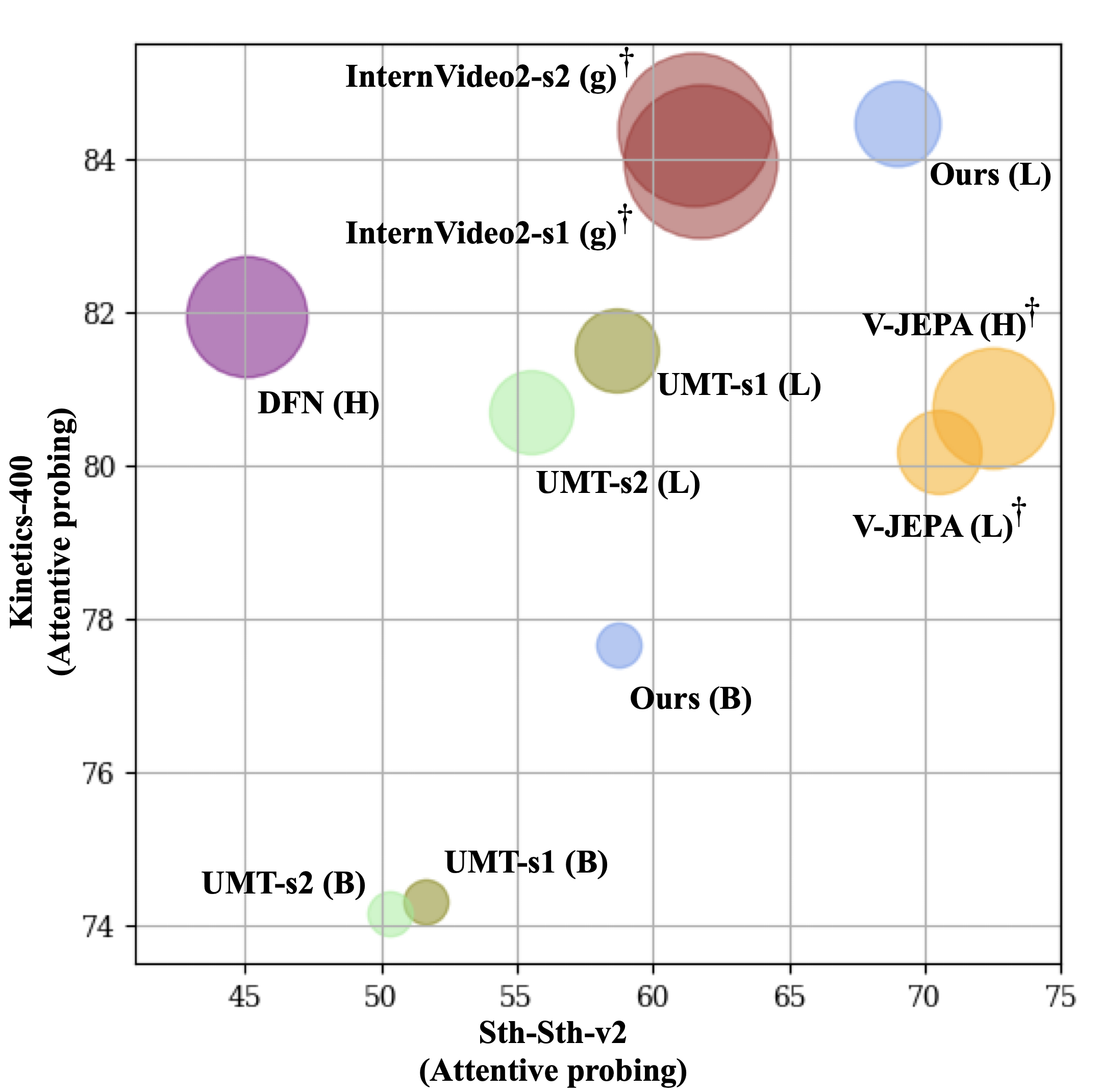}
    \end{subfigure}
    \begin{subfigure}[b]{0.47\textwidth}
        \centering
        \includegraphics[width=\textwidth]{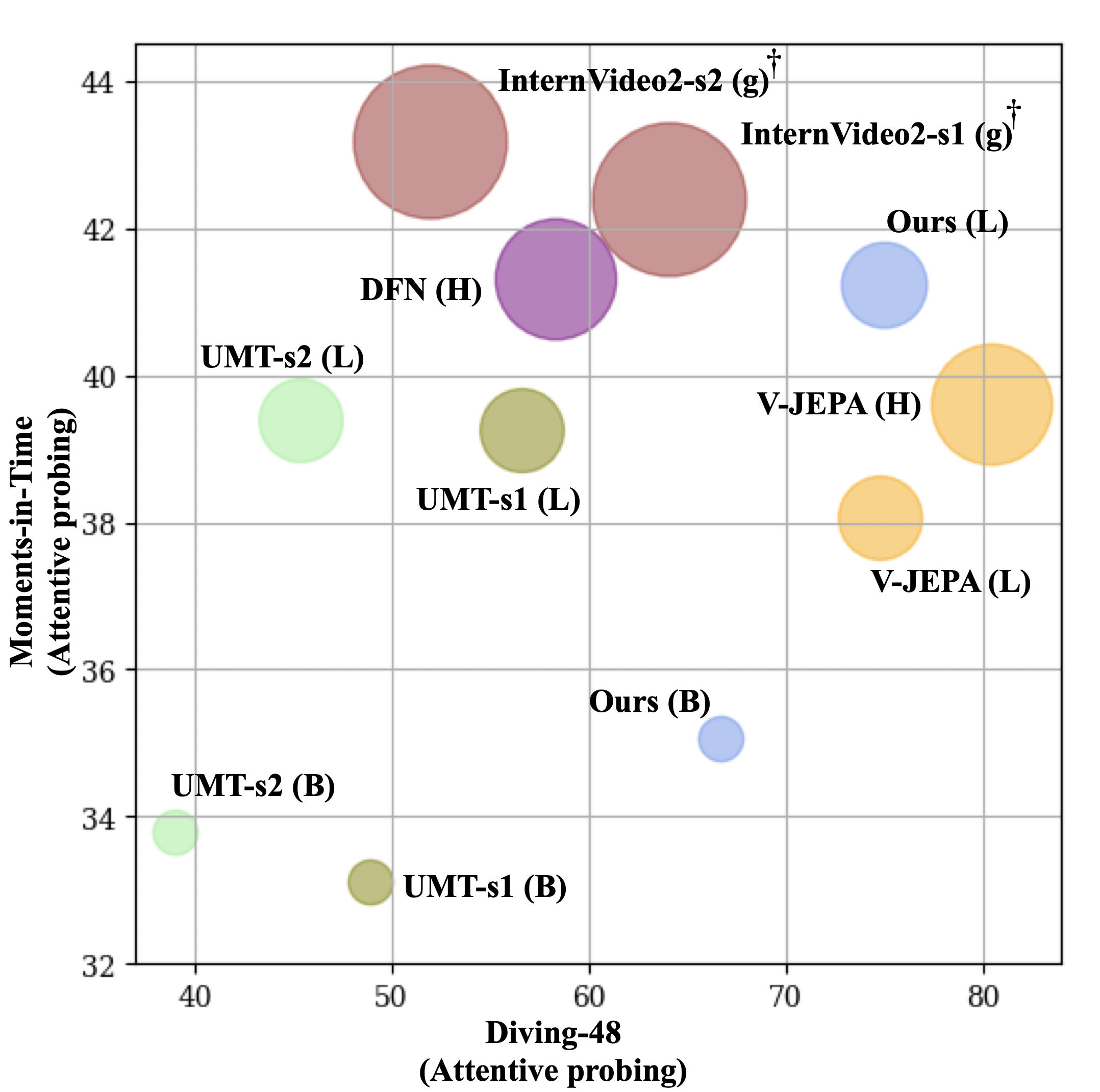}
    \end{subfigure}
    \caption{\textbf{Performance on appearance- vs. motion-centric benchmarks.} Our model can handle both appearance- and motion-centric benchmarks reasonably well. \textsuperscript{\textdagger}~denotes that the pretraining dataset of the model includes the downstream dataset. V-JEPA uses Something-Something-v2 in the pretraining stage. InternVideo2 is pretrained on Moments-in-Time and Something-Something-v2.}

    \label{fig:scatter_plot}
\end{figure}
\begin{figure*}[t!]
    \centering
    \begin{subfigure}[t]{0.24\textwidth}
        \centering
        \includegraphics[width=\linewidth]{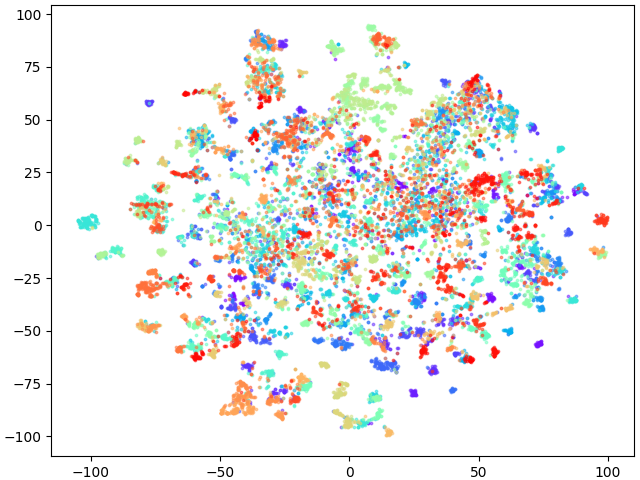}
        \caption{{\ourmodel}}
    \end{subfigure}
    \begin{subfigure}[t]{0.24\textwidth}
        \centering
        \includegraphics[width=\linewidth]{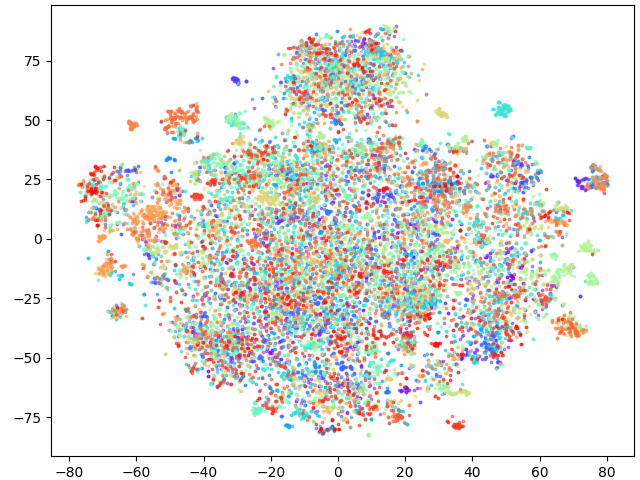}
        \caption{V-JEPA}
    \end{subfigure}
    \tikz{\draw[densely dashed, line width=0.35mm, gray](0,2.5) -- (0,0);}
    \begin{subfigure}[t]{0.24\textwidth}
        \centering
        \includegraphics[width=\linewidth]{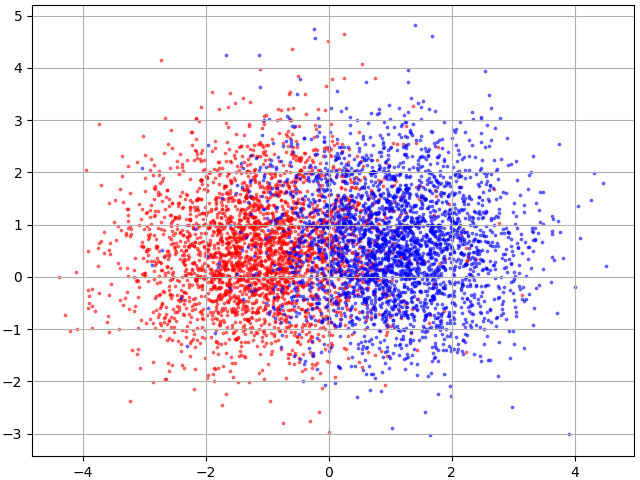}
        \caption{{\ourmodel}}
    \end{subfigure}
    \begin{subfigure}[t]{0.24\textwidth}
        \centering
        \includegraphics[width=\linewidth]{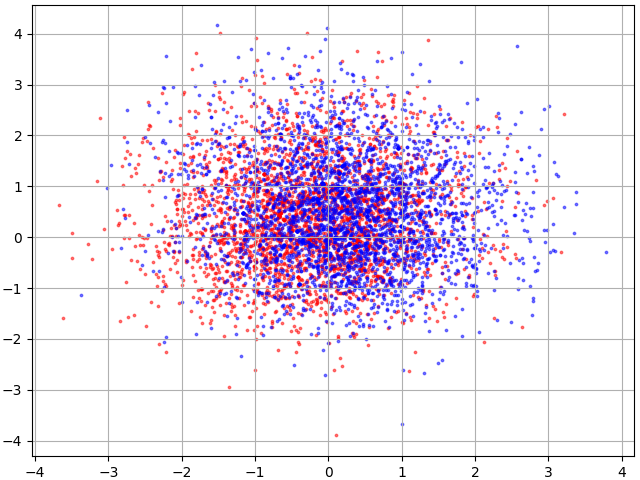}
        \caption{InternVideo2}
    \end{subfigure}
    \caption{(a), (b): t-SNE visualizations of embeddings obtained from the \textbf{K400 validation set} using {\ourmodel} and V-JEPA. (c), (d): LDA visualizations of embeddings from the \textbf{\textcolor{red}{`Moving something up' class}} of the SSv2 validation set and \textbf{\textcolor{blue}{their reversed versions}} using {\ourmodel} and InternVideo2. Details can be found in Figures~\ref{fig:vis_k400}, \ref{fig:updown_down}, \ref{fig:updown_up}, and Section~\ref{sec:embedding_vis}. As seen from (a) and (b), V-JEPA lacks the capability to cluster embeddings of the same class when extracting embeddings from K400, where understanding the visual appearance of each frame is important. From (c) and (d), it is evident that InternVideo2 struggles to distinguish between videos with objects moving in a specific direction and their reversed versions, indicating a limitation in motion understanding capability. In contrast, {\ourmodel} demonstrates both of these capabilities.}
    \label{fig:main_vis}
\end{figure*}

A Foundation Model~(FM) represents a model trained on large-scale data with diverse supervision, possessing the potential to be applied to various tasks within a domain rather than targeting a specific task~\cite{bommasani2021opportunities}.
Recently, Language Foundation Models~(LFMs), represented by Large Language Models~(LLMs), have demonstrated that a single fixed model can effectively handle multiple tasks in general~\cite{achiam2023gpt, touvron2023llama, chiang2023vicuna, jiang2023mistral, chowdhery2023palm}. In the field of computer vision, Image Foundation Models (IFMs) follow this trend by proposing frameworks to encode general information obtainable from images into embedding vectors~\cite{oquab2023dinov2, radford2021clip}. Due to extensive research on large-scale model structures, large datasets, and appropriate training methodologies, fixed general models in the image domain have demonstrated superior or compatible performance across various tasks compared to expert models trained specifically for those tasks~\cite{wang2024internvideo2}.

In the context of video analysis, one possible way to leverage the success of existing IFMs is to treat video as a sequence of images: sampling each frame, embedding them using image encoders, and then aggregating the embeddings~\cite{Luo2021CLIP4Clip}. However, it is known that this method has limitations in capturing the detailed motion in videos~\cite{yuan2023videoglue} and our experiments also demonstrate this tendency, as shown in Figure~\ref{fig:scatter_plot}. Additionally, previous work~\cite{wang2023mvd} observed that the models trained on the images tend to output similar embeddings for the contiguous frames. This approach is akin to dividing the frames into $T$ isolated rooms, having $T$ individuals enter each room to observe their respective frames, and then discussing to compile the information. Such a method makes it difficult to capture the detailed motion of objects within the video and the small changes between frames.

To address this issue, recent works propose utilizing the Vision Transformer~\cite{dosovitskiy2020image} architecture to process all frames simultaneously. These methods can be categorized into two main approaches based on the training methodology. The first approach utilizes distillation, where high-performance IFMs, such as CLIP~\cite{radford2021clip}, are used as a teacher for distillation~\cite{li2023umt,wang2024internvideo2}. The second approach is based on masked modeling, where the model is trained to predict the missing information of inputs from the given partial input~\cite{tong2022videomae,wang2023videomae2,feichtenhofer2022stmae,bardes2024vjepa}. Since a video is essentially a sequence of images, it is crucial to understand the \textbf{appearance} of objects in the given data. Unlike images, videos also involve the time axis, making it essential to understand the \textbf{motion} of the objects. However, as shown in Figure~\ref{fig:scatter_plot} and Figure~\ref{fig:main_vis}, distillation-based methods such as UMT and InternVideo2 struggle with motion-sensitive benchmarks such as Something-Something-v2~\cite{benchmark_ssv2} and Diving-48~\cite{benchmark_dv48}. On the other hand, masked modeling-based methods (e.g., V-JEPA) underperform on appearance-centric benchmarks such as Kinetics-400~\cite{benchmark_k400} and Moments-in-Time~\cite{benchmark_mit}.

In this work, we introduce \textbf{\ourmodel}, a model that can provide an embedding vector for a video, capturing both appearance and motion. 
As shown in Figure~\ref{fig:main_plot}, although {\ourmodel} is trained solely with publicly available datasets described in Table~\ref{tab:pretrain}, demonstrates noticeable performance on both appearance- and motion-centric action recognition benchmark datasets. 
Furthermore, in addition to the action recognition task, {\ourmodel} achieves state-of-the-art performance on various video-centric tasks such as temporal action localization, spatiotemporal action localization, and temporal action segmentation, showing its strong spatial and temporal understanding capabilities.

To conduct a detailed analysis of various VFMs, including {\ourmodel}, we utilize several commonly used evaluation and analysis methods~\cite{yuan2023videoglue, bardes2024vjepa}. However, given that these existing methods are insufficient for a comprehensive analysis of VFMs, we have improved some of them and proposed new analytical approaches. For example, in Section~\ref{sec:forward_reverse}, we validate whether a VFM can distinguish videos based solely on the direction of motion, independent of appearance. This is achieved by visualizing the embeddings of original videos and their reversed versions to determine if their embedding distributions are separable.

In this technical report, we not only introduce {\ourmodel} but also highlight several important perspectives and evaluation methods that are essential for advancing the field of video understanding. Additionally, we propose key directions for future research to further the development of this area.
\section{{\ourmodel} \& Video Foundation Model Evaluation Framework}

\begin{table}[t]
\centering
\caption{\textbf{Summary of datasets used for pretraining {\ourmodel}.}}
\begin{tabular}{lcc}
\hline
Pretraining Dataset & Type  & \# of Clips (Images) \\ \hline
Kinetics 710        & Video & 658K                 \\
Howto360K           & Video & 360K                 \\
WebVid10M           & Video & 10.73M               \\
COCO                & Image & 113K                 \\
SBU Captions        & Image & 860K                 \\
Visual Genome       & Image & 100K                 \\
CC3M                & Image & 2.88M                \\
CC12M               & Image & 11.00M               \\ \hline
\end{tabular}
\label{tab:pretrain}
\end{table}
\begin{table}[t]
\centering
\caption{\textbf{Summary of benchmarks used for evaluation.}}
\label{tab:eval_benchmark} 
\resizebox{\linewidth}{!}{
\begin{tabular}{l | l r l l l}
\hline
Task  & Dataset & \# Videos (train / val) & Clip Length   & Domain & Note   \\ \hline

\multirow{5}{*}{AR}  & Kinetics 400  & 235,693 / 19,165  & 10 seconds & Web & Appearance   \\

   & MiT  & 791,246 / 33,898  & 3 seconds  & Web & Appearance   \\
   & SthSth-v2 & 168,913 / 24,777  & 2-6 seconds   & Crowd-source & Motion \\
   & Diving 48  & 15,027 / 1,970 & 5 seconds  & Web & Motion \\
   & Epic Kitchens & 67,217 / 9,668 & $\sim$10 seconds & Crowd-source & Ego-centric  \\ \hline
   
\multirow{2}{*}{TAL} & ActivityNet v1.3 & 10,024 / 4,926 & 5-10 minutes  & Web & Temporal  \\
   & THUMOS14   & 200 / 212   & 4 minutes  & Web & Temporal  \\ \hline
STAL  & AVA v2.2   & 210,634 / 57,371  & 15 minutes & Movie  & Spatio-temporal \\ \hline
\multirow{3}{*}{TAS} & 
   50Salads & 40 / 10 & 5 minutes  & Web & Temporal \\
   & GTEA & 21 / 7 & 1-3 minutes  & Web & Ego-centric \& Temporal \\ 
   & Breakfast & $\sim$1,284 / $\sim$428 & 2-3 minutes  & Web & Temporal \\ \hline
\end{tabular}
}
\end{table}







In this section, we analyze VAM's feature space compared to previous Video Foundation Models (VFMs). For this purpose, we focus on four representative tasks: Action Recognition (AR), Temporal Action Localization (TAL), Spatio-temporal Action Localization (STAL), and Temporal Action Segmentation (TAS). We evaluate the performance of each VFM across different settings, benchmarks, and evaluation methods. Details of the benchmarks used for evaluation are provided in Table~\ref{tab:eval_benchmark}. We compare our model with recent state-of-the-art VFMs, including Unmasked Teacher (UMT)~\cite{li2023umt}, V-JEPA~\cite{bardes2024vjepa}, and InternVideo2~\cite{wang2024internvideo2}. Additionally, we include the Data Filtering Network (DFN)~\cite{fang2023dfn} as an appearance-centric model that processes video frame by frame, rather than handling it as a whole. To extract the embedding of a video, we apply a mean-pooling operation to the frame-wise embeddings. This strategy can be seen as a way to utilize an Image Foundation Model (IFM) in the video domain.

\subsection{Training Details and Frame Sampling}
\label{subsec:training_details_clip_sampling}
\noindent\textbf{Architecture.} We adopt a Vision Transformer~(ViT)~\cite{dosovitskiy2020image} architecture, specifically ViT-B (Base, 86M parameters) and ViT-L (Large, 307M parameters). An input video is tokenized into multiple patches and then processed through the transformer to produce patch-wise embeddings, which are subsequently pooled to obtain the overall embedding of the input.

\noindent\textbf{Pretrain Dataset.} For pretraining, we use Kinetics-710~\cite{li2022uniformerv2}, HowTo360K (a subset of HowTo100M~\cite{miech2019howto100m}), WebVid10M~\cite{bain2021frozen}, and a mixture of 15M publicly available image datasets to enhance image understanding capabilities. Details of the pretraining datasets can be found in Table~\ref{tab:pretrain}.

\noindent\textbf{Training Objective.} Regarding the objective function, we leverage the strengths of existing distillation-based and masked modeling-based methods. Using the masked modeling schema as the baseline, we diversify the reconstruction targets to find the optimal objective for training a robust model that excels in both motion and appearance understanding. We train our model from scratch with this objective and the aforementioned datasets.

\begin{figure*}[t!]
    \centering
    \includegraphics[width=0.80\linewidth]{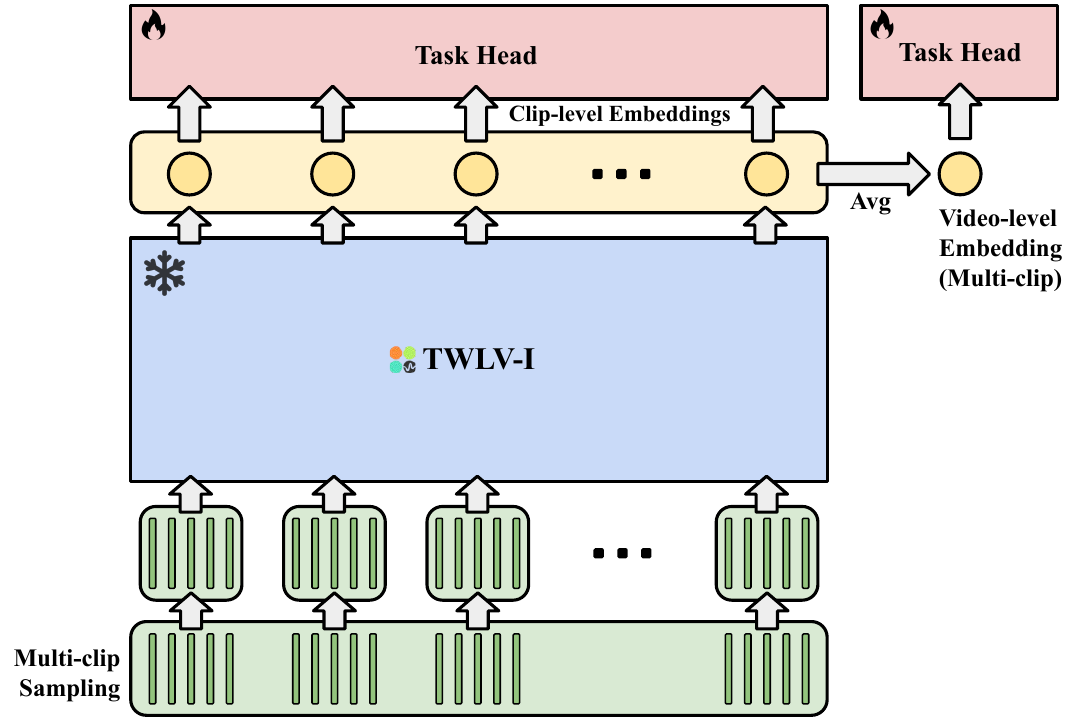}
    \caption{\textbf{Overall architecture of the evaluation framework including TWLV-I.} In \texttt{Multi-Clip Embedding}, the video is divided into multiple clips, and an embedding is produced from each clip. These clip-level embeddings are either all passed to the downstream task or averaged before being passed to the task head.}
    \label{fig:architecture}
\end{figure*}

\noindent\textbf{Frame Sampling.} Because the computational complexity of the ViT architecture increases quadratically with the number of tokens, there is a constraint on the maximum number of frames that can be processed at once. Therefore, frames of an input video must be sub-sampled before being provided to the model. The most straightforward method is to sample the $N$ frames with a uniform stride~(at fixed intervals), regardless of the video's length (\texttt{Uniform Embedding}). However, as the video length increases, the time intervals between frames become longer, potentially losing temporal coherence between frames. To address this, for specific tasks, instead of \texttt{Uniform Embedding}, we fix the time length of a clip to $T$ seconds and divide the video into $M$ clips of $T$ seconds each. Then, $N$ frames are sampled from each clip to obtain $M$ embeddings (\texttt{Multi-Clip Embedding}). In this setting, the number of embeddings increases as the video lengthens. When representing the entire video with a single embedding, we average the $M$ embeddings. Research on finding better methods than averaging to aggregate $M$ embeddings into a single video embedding could be an interesting future research topic. The differences between the overall evaluation pipeline and the two frame sampling methods are shown in Figure~\ref{fig:architecture}.

\subsection{Action Recognition}
\label{sec:action_recognition}
Action Recognition~(AR) is a video classification task that aims to classify videos into predefined human action categories. For the action recognition task, we adopt five representative benchmarks: Kinetics-400~(K400)~\cite{benchmark_k400}, Something-Something-v2~(SSv2)~\cite{benchmark_ssv2}, Moments-in-Time~(MiT)~\cite{benchmark_mit}, Diving-48~(DV48)~\cite{benchmark_dv48}, and Epic-Kitchens~(EK)~\cite{benchmark_ek}. We note that K400 and MiT are known to be appearance-focused, while SSv2 and DV48 are motion-centric datasets~\cite{yuan2023videoglue,bardes2024vjepa}. We utilize the multi-view classification method commonly used in action recognition tasks. Specifically, we first spatially resize the input video so that its short side length matches the input resolution. Then, we uniformly sample $m$ clips along the long side direction and $n$ clips at equal intervals in the temporal direction. This results in a total of $m \times n$ clips. We then calculate the class probabilities for each of these clips and obtain the final output by averaging these probabilities.

\subsubsection{Linear Probing}
\begin{table}
\centering
 \caption{\textbf{Linear probing on recognition benchmarks.} \textsuperscript{\textdagger} denotes that the pretraining dataset for the model includes the downstream dataset. We note that all the pretraining dataset of video models includes Kinetics-400 (K400).}
 \resizebox{\linewidth}{!}{
 \begin{tabular}{l| c | c c c c c c c c c c | c |c c}
  \hline

\multirow{2}{*}{Method} & \multirow{2}{*}{Arch.} & \multicolumn{2}{c}{K400} & \multicolumn{2}{c}{MiT} & \multicolumn{2}{c}{SSv2} & \multicolumn{2}{c}{DV48} & \multicolumn{2}{c}{EK} & {Avg.} & \multicolumn{2}{c}{IN-1K} \\
  & & Top1 & Top5 & Top1 & Top5 & Top1 & Top5 & Top1 & Top5 & Top1 & Top5 & Top1 & Top1 & Top5 \\
  
  \hline
UMT$_{s1}$   & ViT-B & 66.17  & 86.25  & 26.87  & 51.60 & 17.46  & 47.92  & 17.46  & 47.92  & 35.38  & 77.58 & 32.67  & 70.84  & 90.55  \\
UMT$_{s2}$   & ViT-B & 69.47  & 88.65  & 29.79  & 56.17 & 18.22  & 48.63  & 18.22  & 48.63  & 37.41  & 76.67 & 34.62  & \textbf{74.75}  &\textbf{ 93.17}  \\
\rowcolor{tl!50} \adjustbox{valign=c}{\includegraphics[width=0.02\linewidth]{tables/logo.png}} {\ourmodel}& ViT-B & \textbf{74.17}  & \textbf{91.52}& \textbf{30.59}  & \textbf{56.89} & \textbf{37.19}  & \textbf{66.79}  & \textbf{30.41}& \textbf{67.82}  & \textbf{42.05}  & \textbf{80.03} & \textbf{42.88} & 61.14  & 83.95  \\ \hline
UMT$_{s1}$   & ViT-L & 74.50  & 91.35  & 32.87  & 60.02 & 24.74  & 52.11  & 22.69  & 55.23  & 37.37  & 76.39 & 38.43  & 76.05  & 93.47  \\
UMT$_{s2}$   & ViT-L & 76.60  & 92.48  & 35.69  & 63.71 & 31.02  & 61.25  & 21.42  & 54.92  & 39.81  & 79.28 & 40.91 & \textbf{79.24}  & \textbf{95.12}  \\
V-JEPA & ViT-L & 70.77  & 90.00  & 28.83  & 54.41 & \textbf{52.50\textsuperscript{\textdagger}}  & \textbf{80.56\textsuperscript{\textdagger}}  & 22.86  & 56.84  & 45.34  & 77.07 & 44.06 & 56.67  & 80.37  \\

\rowcolor{tl!50} \adjustbox{valign=c}{\includegraphics[width=0.02\linewidth]{tables/logo.png}} {\ourmodel}& ViT-L & \textbf{80.17}  & \textbf{92.83}  & \textbf{36.29}  & \textbf{64.79} & {46.41}  & {75.79}  & \textbf{32.84}  & \textbf{70.36}  & \textbf{47.34}  & \textbf{80.14} & \textbf{48.61} & 72.98  & 92.13  \\ \hline

 DFN & ViT-H & 77.99  & 92.62  & {37.27}  & {65.21} & 26.28  & 54.52  & {27.36}  & {64.62}  & 38.07  & 77.32 & 41.39  & {84.55}  & {97.12}  \\
 V-JEPA & ViT-H & 70.75	& 89.36 & 28.95 &	54.19 & 53.41\textsuperscript{\textdagger} &	81.18\textsuperscript{\textdagger} & 29.34 &	67.82 & 47.30 &	78.40 & 45.95 & 58.26	& 80.96 \\
InternVideo2$_{s1}$ & ViT-g & 78.20  & 92.97  & 38.47\textsuperscript{\textdagger}  & 66.84\textsuperscript{\textdagger} & 26.18\textsuperscript{\textdagger}  & 54.72\textsuperscript{\textdagger}  & 25.13  & 59.34  & 37.09  & 78.59 & 41.01 & 79.78  & 95.31  \\
 InternVideo2$_{s2}$ & ViT-g & {81.81}  & {94.61}  & {40.52\textsuperscript{\textdagger}}  & {69.33\textsuperscript{\textdagger}} & 40.13\textsuperscript{\textdagger}  & 72.30\textsuperscript{\textdagger}  & 23.71  & 54.82  & {42.69}  & {81.03} & 45.77 & {80.06}  & {95.38}  \\


  \hline \hline

\rowcolor{tl!20} \adjustbox{valign=c}{\includegraphics[width=0.02\linewidth]{tables/logo.png}} {\ourmodel}~(+SSv2)   & ViT-L & 80.73   & 94.47  & 37.19 & 65.37 & 48.14\textsuperscript{\textdagger}   & 78.67\textsuperscript{\textdagger}   & 31.07   & 71.52   & 46.58   & 81.97 & 48.74  & 73.64   & 92.54   \\

  \hline 
 \end{tabular}
}

 \label{tab:ac_linprob}
\end{table}


\noindent\textbf{Evaluation Settings.} To evaluate the action recognition capability in terms of the clip-wise embeddings, we first adopt linear probing, which involves freezing the feature extractor (i.e., backbone model) and training a linear classifier on top of it. The linear classifier consists of a weight matrix that matches the embedding vector dimension to the number of classes. Details of the hyperparameters are provided in Table~\ref{tab:AP_hyperparams}.

\noindent\textbf{Results and Discussion.} As shown in Table~\ref{tab:ac_linprob}, UMT and InternVideo2 demonstrate limitations on motion-centric benchmarks such as SSv2 and DV48, which alignes with the discussion in Section~\ref{sec:intro}. Similarly, DFN, being an image encoder, lacks the capability to effectively understand these benchmarks.
In contrast, V-JEPA shows superior performance on motion-centric datasets. However, it exhibits limitations on appearance-centric benchmarks such as K400 and MiT. While the V-JEPA model at the ViT-H scale performs better than the ViT-L scale on SSv2, DV48, and EK, its performance either improves very slightly or even decreases on appearance-centric benchmarks (K400 and MiT). This indicates that scaling up V-JEPA enhances motion understanding capability but does not improve its already limited appearance capability. Unlike its competitors, our model generally performs well across all benchmarks. Notably, {\ourmodel} achieves the best performance on almost all benchmarks within the same architecture scale~(both ViT-B and ViT-L). For the SSv2 dataset, our model outperforms even the larger-scale models like ViT-H~(DFN) and ViT-g~(InternVideo2), except for V-JEPA, which uses SSv2 in the pretraining stage. Furthermore, our ViT-L model surpasses other larger-scale models on the EK and DV48 benchmarks.

\subsubsection{Attentive Probing}

\begin{table}
\centering
 \caption{\textbf{Hyperparameters for linear probing (LP) and attentive probing (AP) evaluation.} The hyperparameters are from V-JEPA~\cite{bardes2024vjepa} with minor changes.}
 \resizebox{\linewidth}{!}{
 \begin{tabular}{l c c c c  c c  c c c c c c}
 \toprule
 Hyper-parameter & \multicolumn{2}{c}{K400} & \multicolumn{2}{c}{MiT} & \multicolumn{2}{c}{SSv2} & \multicolumn{2}{c}{DV48} & \multicolumn{2}{c}{EK} & \multicolumn{2}{c}{IN-1K} \\ \hline
 - & LP & AP & LP & AP & LP & AP & LP & AP & LP & AP & LP & AP \\
 train\_num\_clips & \multicolumn{12}{c}{1} \\
 views ($m \times n$) & \multicolumn{2}{c}{3 $\times$ 4} & \multicolumn{2}{c}{3 $\times$ 4} & \multicolumn{2}{c}{3 $\times$ 1} & \multicolumn{2}{c}{3 $\times$ 4} & \multicolumn{2}{c}{3 $\times$ 4} & \multicolumn{2}{c}{-} \\
 num\_frames & \multicolumn{2}{c}{16} & \multicolumn{2}{c}{16} & \multicolumn{2}{c}{16} & \multicolumn{2}{c}{16} & \multicolumn{2}{c}{16} & \multicolumn{2}{c}{-} \\ 
 temporal\_stride & \multicolumn{2}{c}{4} & \multicolumn{2}{c}{1} & \multicolumn{2}{c}{1} & \multicolumn{2}{c}{4} & \multicolumn{2}{c}{1} &  \multicolumn{2}{c}{-}\\ 
 horizontal\_flip & \multicolumn{2}{c}{True} & \multicolumn{2}{c}{False} & \multicolumn{2}{c}{False} & \multicolumn{2}{c}{False} & \multicolumn{2}{c}{True} & \multicolumn{2}{c}{True} \\ 
 random\_resize & \multicolumn{12}{c}{(0.3, 1.0)} \\ 
 aspect\_ratio & \scriptsize{0.75, 1.33} & \scriptsize{0.5, 2.0} & \scriptsize{0.75, 1.33} & \scriptsize{0.5, 2.0} & \scriptsize{0.75, 1.33} & \scriptsize{0.5, 2.0} & \scriptsize{0.75, 1.33} & \scriptsize{0.5, 2.0} & \scriptsize{0.75, 1.33} & \scriptsize{0.5, 2.0} & \scriptsize{0.75, 1.33} & \scriptsize{0.5, 2.0} \\ 
 \hline
 batch\_size & 1024 & 256 & 1024 & 256 & 1024 & 256 & 1024 & 256 & 1024 & 256 & 4096 & 512\\ 
 epochs & 150 & 20 & 50 & 20 & 50 & 20 & 300 & 50 & 150 & 50 & 90 & 20 \\ 
 warmup & 10 & 0 & 10 & 0 & 10 & 0 & 10 & 0 & 10 & 0 & 0 & 0\\
 scheduler & \multicolumn{12}{c}{cosine decay} \\
 lr & 0.1 & 0.001 & 0.1 & 0.001 & 0.075 & 0.001 & 0.02 & 0.001 & 0.1 & 0.001 & 0.1 & 0.001 \\ 
 final\_lr & \multicolumn{12}{c}{0.0} \\ 
 weight\_decay & 0 & 1e-5 & 0 & 1e-5 & 0 & 1e-5 & 0 & 1e-5 & 0 & 1e-5 & 0 & 0.001 \\

 \bottomrule
 
 \end{tabular}
}

 \label{tab:AP_hyperparams}
\end{table}
\begin{table}
 \centering
  \caption{\textbf{Attentive probing on recognition benchmarks}. \textsuperscript{\textdagger} denotes that the pretraining dataset for the model includes the downstream dataset. $^*$ indicates that the results are from V-JEPA~\cite{bardes2024vjepa}.
  The accuracy of VideoPrism is from the original paper. We note that all the pretraining dataset of video models includes Kinetics-400 (K400).}
\resizebox{\linewidth}{!}{
 \begin{tabular}{l| c | c c c c c c c c c c | c |c c}
  \hline

\multirow{2}{*}{Method} & \multirow{2}{*}{Arch.} & \multicolumn{2}{c}{K400} & \multicolumn{2}{c}{MiT} & \multicolumn{2}{c}{SSv2} & \multicolumn{2}{c}{DV48} & \multicolumn{2}{c}{EK} & {Avg.} & \multicolumn{2}{c}{IN-1K} \\
  & & Top1 & Top5 & Top1 & Top5 & Top1 & Top5 & Top1 & Top5 & Top1 & Top5 & Top1 & Top1 & Top5 \\
  \hline
UMT$_{s1}$   & ViT-B & 74.32   & 91.46   & 33.12   & 60.81 & 51.64   & 80.03   & 48.88   & 86.60   & 54.76   & 87.04 & 52.54   & 78.16   & 94.75   \\
UMT$_{s2}$   & ViT-B & 74.15   & 91.31   & 33.80   & 62.12 & 50.30   & 79.30   & 38.93   & 77.82   & 52.74   & 87.16 & 49.98   & \textbf{79.83}   & \textbf{95.73 }  \\
\rowcolor{tl!50} \adjustbox{valign=c}{\includegraphics[width=0.02\linewidth]{tables/logo.png}} {\ourmodel}  & ViT-B & \textbf{77.66}   & \textbf{93.47}   & \textbf{35.06}   & \textbf{63.26} & \textbf{58.74}   & \textbf{85.69}   & \textbf{66.70}   & \textbf{93.10}   & \textbf{58.31}   & \textbf{88.50} & \textbf{59.29}   & 70.27 & 90.72 \\ \hline
OmniMAE$^*$ & ViT-L & 65.6 & - & - & - & 60.6 & - & - & - & - & - & - & 75.1 & - \\
VideoMAE$^*$ & ViT-L & 77.8 & - & - & - & 65.5 & - & - & - & - & - & - & 71.1 & - \\
Hiera$^*$ & Hiera-L & 75.5 & - & - & - & 64.2 & - & - & - & - & - & - & 68.9 & - \\
MVD$^*$ & ViT-L & 79.4 & - & - & - & 66.5 & - & - & - & - & - & - & 73.3 & - \\
UMT$_{s1}$   & ViT-L & 81.50   & 95.03   & 39.25   & 67.58 & 58.68   & 85.13   & 56.60   & 89.39   & 57.91   & 88.32  & 58.79 & 82.64   & 96.77   \\
UMT$_{s2}$   & ViT-L & 80.70   & 94.81   & 39.39   & 68.69 & 55.53   & 83.56   & 45.38   & 84.35   & 55.70   & 88.12  & 55.34 & \textbf{83.71}   & \textbf{97.17}   \\
V-JEPA  & ViT-L & 80.19   & 94.63   & 38.06   & 67.02 & \textbf{70.53\textsuperscript{\textdagger}}   & \textbf{92.97\textsuperscript{\textdagger}}   & {74.72}   & {95.58}   & {62.34}   & {89.62} & 65.17  & 73.60   & 91.92   \\
\rowcolor{tl!50} \adjustbox{valign=c}{\includegraphics[width=0.02\linewidth]{tables/logo.png}} {\ourmodel}  & ViT-L & \textbf{84.46}   & \textbf{96.58}   & \textbf{41.23}   & \textbf{70.45} & 68.95   & 91.84   & \textbf{74.98}   & \textbf{96.29}   & \textbf{63.70}   & \textbf{90.16} & \textbf{66.66}  & 79.19   & 95.26   \\ \hline

DFN   & ViT-H & 81.96   & 95.15   & {41.30}   & 70.06 & 45.03   & 76.31   & 62.18   & 93.30   & 51.85   & 87.84 & 56.46   & 87.54   & 98.32   \\
V-JEPA & ViT-H & 80.75	& 95.03 & 39.60	& 68.18 & 72.48\textsuperscript{\textdagger}	& 93.77\textsuperscript{\textdagger} & 80.41	& 97.97 & 63.94	& 89.57 & 67.44 & 73.49	& 91.66 \\

InternVideo2$_{s1}$ & ViT-g & 83.96   & 95.97   & 42.39\textsuperscript{\textdagger}   & 71.35\textsuperscript{\textdagger} & 61.76\textsuperscript{\textdagger}   & 86.77\textsuperscript{\textdagger}   & 64.01   & 91.73   & 59.04   & 89.27 & 62.23  & 85.77   & 97.98   \\
InternVideo2$_{s2}$ & ViT-g & {84.39}   & {96.27}   & {43.18\textsuperscript{\textdagger}}   & {72.52\textsuperscript{\textdagger}} & 61.54\textsuperscript{\textdagger}   & 87.77\textsuperscript{\textdagger}   & 51.88   & 85.18   & 56.87   & 89.36 & 59.57   & {85.28}   & {97.75}   \\
\textcolor{gray}{VideoPrism} & \textcolor{gray}{ViT-g} & \textcolor{gray}{87.2} & \textcolor{gray}{-} & \textcolor{gray}{45.5} & \textcolor{gray}{-} & \textcolor{gray}{68.5} & \textcolor{gray}{-} & \textcolor{gray}{71.3} & \textcolor{gray}{-} & \textcolor{gray}{-} & \textcolor{gray}{-} & \textcolor{gray}{-} & \textcolor{gray}{-} & \textcolor{gray}{-} \\


  \hline \hline

\rowcolor{tl!20} \adjustbox{valign=c}{\includegraphics[width=0.02\linewidth]{tables/logo.png}} {\ourmodel}~(+SSv2)   & ViT-L & 84.54   & 96.43   & 42.00   & 71.38 & 71.12\textsuperscript{\textdagger}   & 93.02\textsuperscript{\textdagger}   & 74.57   & 96.50   & 65.17   & 90.35  & 66.88   & 80.84   & 95.88   \\

\hline
  
 \end{tabular}
}

 \label{tab:ac_attnprob}
\end{table}

\noindent\textbf{Evaluation Settings.} While linear probing can evaluate the quality of the clip-wise embeddings, it cannot fully reflect the models' capabilities, especially since models trained with patch-level supervision lack the frame-level descriptor~\cite{el2024aim}. To address this, we also validate the models using attentive probing. This method involves training a single attention layer with a learnable class token on top of the frozen models. The output class token is then passed through a linear classifier, and we measure the top-1 accuracy. We follow the same training schedule as V-JEPA~\cite{bardes2024vjepa}, with detailed training parameters provided in Table~\ref{tab:AP_hyperparams}.

\noindent\textbf{Results and Discussion.} As shown in Table~\ref{tab:ac_attnprob}, the attentive probing results confirm that our model achieves superior performance across both appearance- and motion-centric benchmarks compared to other models. Moreover, these results provide additional insights not captured by the linear probing results, such as the observation that the performance of some models in stage 2 is lower than in stage 1. Specifically, for UMT, a performance drop occurs in all benchmarks except MiT. For InternVideo2, the drop is observed in all benchmarks except K400 and MiT. Both UMT and InternVideo2 are trained with patch-level loss in stage 1, while state 2 involves aligning embedding vectors with other domains. This indicates that stage 2 weakens the patch-wise representation, which is particularly detrimental for motion-centric datasets, as evidenced by the significant performance drop in DV48. Our model shows even better performance in attentive probing than in linear probing, outperforming larger models like InternVideo2 not only on motion-centric benchmarks but also on appearance-centric ones. This suggests that our model has a richer patch-wise representation capability compared to other models. For the last row of VideoPrism~\cite{zhao2024videoprism}, the reported numbers are taken directly from the corresponding paper. It is important to note that we follow V-JEPA's attentive probing schedule, which differs from the training schedule of VideoPrism.

\begin{table}
 \centering
  \caption{\textbf{K-Nearest Neighbors}. \textsuperscript{\textdagger} denotes that the pretraining dataset for the model includes the downstream dataset. Note that all the pretraining dataset of video models includes Kinetics-400 (K400).}
\resizebox{0.7\linewidth}{!}{
 \begin{tabular}{l| c | c c c | c c c}
  \hline 
  &  & \multicolumn{3}{c}{K400} & \multicolumn{3}{c}{SSv2} \\ 
  Method & Arch. & Uniform & Clip & Video & Uniform & Clip & Video \\ \hline
UMT$_{s2}$ & ViT-B & 50.95&	55.76	&54.64&	12.00&	12.46&	11.92 \\
\rowcolor{tl!50} \adjustbox{valign=c}{\includegraphics[width=0.02\linewidth]{tables/logo.png}} {\ourmodel} & ViT-B & \textbf{57.51} &	\textbf{63.64}&	\textbf{62.59}&	\textbf{19.82}&	\textbf{19.67}&	\textbf{18.48}\\ \hline
UMT$_{s2}$ & ViT-L & 60.68 & 	64.68 &	63.76 &	14.47 &	14.76 &	14.24 \\
V-JEPA & ViT-L & 39.95 &47.62&	46.00&	\textbf{25.82}\textsuperscript{\textdagger}&	\textbf{25.93}\textsuperscript{\textdagger}&	\textbf{23.23}\textsuperscript{\textdagger} \\
\rowcolor{tl!50} \adjustbox{valign=c}{\includegraphics[width=0.02\linewidth]{tables/logo.png}} {\ourmodel} & ViT-L & \textbf{65.97} &	\textbf{70.82} &	\textbf{69.88} &	19.47 &	19.73 &	19.11 \\
\hline
DFN & ViT-H & 63.78 &	68.07 &	66.97 &	15.30 &	16.21 &	15.62 \\
V-JEPA & ViT-H & 34.46 &	42.45	& 40.68 &	24.67\textsuperscript{\textdagger} &	24.90\textsuperscript{\textdagger} &	21.54\textsuperscript{\textdagger} \\
InternVideo2 & ViT-g & 76.15 &	79.43 &	78.46 &	22.65 &	23.38 &	22.86 \\
\hline \hline
\rowcolor{tl!20} \adjustbox{valign=c}{\includegraphics[width=0.02\linewidth]{tables/logo.png}} {\ourmodel}~(+SSv2) & ViT-L & 67.83 &	72.14 &	71.21 &	21.99 &	22.33 &	21.94 \\

\hline
  
 \end{tabular}
}

 \label{tab:knn}
\end{table}

\subsubsection{K-Nearest Neighbors}

\noindent\textbf{Evaluation Settings.} In the cases of linear probing and attentive probing, the optimization process uses stochastic gradient descent, which can result in varying performances due to differences in initialization and batch order. 
Additionally, as the embedding vector dimension increases, the number of learnable parameters also increases, potentially making the evaluation process unfair. Therefore, it is necessary to compare the quality of embedding vectors in a parameter-free manner. To this end, we adopt the K-Nearest Neighbors~(KNN) classification task, one of the simplest and most representative non-parametric methods. We obtain two versions of embeddings through the methods introduced in Section~\ref{subsec:training_details_clip_sampling}: \texttt{Uniform Embedding} (`Uniform' in Table~\ref{tab:knn}) and \texttt{Multi-Clip Embedding}. For \texttt{Multi-Clip Embedding}, we fix the length of each clip to 2 seconds. Unlike \texttt{Uniform Embedding}, \texttt{Multi-Clip Embedding} can yield more than one vector per video. In this case, there are two evaluation methods. One method is to average all the embeddings to create a single video embedding (`Video' in Table~\ref{tab:knn}). The other method, similar to the multi-view classification introduced in Section~\ref{sec:action_recognition}, sums the votes received by each clip to determine the final class (`Clip' in Table~\ref{tab:knn}). We conduct three types of evaluations in total. The results are presented in Table~\ref{tab:knn}.

\noindent\textbf{Results and Discussion.} Similar to the linear probing and attentive probing results, our model demonstrates generally good performance on both K400 and SSv2 compared to the models of the same scale. However, unlike the attentive probing results, our model performs worse than InternVideo2 on K400 and SSv2. This indicates that there is room for improvement in utilizing embeddings extracted from our model in a non-parametric manner. An interesting point observed in Table~\ref{tab:knn} is the difference between uniform sampling and clip sampling. Since K400 and SSv2 mostly consist of videos longer than 2 seconds, multiple clips are obtained from most videos. As a result, the `Clip' performance, which utilizes each clip independently, shows the highest accuracy. The tendencies differ in the `Uniform' and `Video' columns. K400 videos are generally longer (around 10 seconds) than SSv2 videos, making them more vulnerable to uniform sampling (longer intervals between frames). However, as K400 videos are more appearance-centric, the information loss caused by averaging clip embeddings is minimal. In contrast, SSv2 videos are relatively shorter, resulting in a smaller performance drop due to uniform sampling. Since SSv2 contains many temporally dynamic videos, simply averaging clip embeddings does not boost and may even hurt performance compared to uniform sampling. This experiment highlights the need for further research and consideration on representing embeddings for videos longer than a single clip length, and even for videos lasting several hours.

\subsubsection{Pretraining with Something-Something-v2 Dataset.}
\noindent\textbf{Evaluation Settings.} Unlike V-JEPA, we exclude SSv2 from our pretraining dataset. Because SSv2 contains classes such as `Moving something up' and `Moving something down' that cannot be distinguished by appearance alone from a single frame, we consider it better to exclude SSv2 from pretraining to precisely measure the motion understanding capability of the models. However, for a fair comparison with V-JEPA and to understand the impact of including SSv2 in the pretraining dataset, we also train an additional model that includes SSv2 in the pretraining phase. The results of our model with SSv2 are presented at the bottom of Table~\ref{tab:ac_linprob}, Table~\ref{tab:ac_attnprob}, and Table~\ref{tab:knn}.

\noindent\textbf{Results and Discussion.} 
Including SSv2 in the pretraining stage results in improved performance on SSv2 across linear probing, attentive probing, and KNN evaluations. Additionally, performance generally increases on other benchmarks as well. This gain can be attributed to the increased data scale used during pretraining. Notably, in attentive probing, our model trained with SSv2 outperforms V-JEPA with the same architecture, achieving the best performance among all competitors. Despite these improvements, the KNN performance is still not as strong as that of V-JEPA and InternVideo2. This indicates that our model has a relatively weaker video-level representation compared to its patch-level representation. Addressing this imbalance should be one of the directions for future enhancements.

\subsection{ImageNet Classification}
\label{sec:imagenet}
To validate whether the foundation models trained on the video dataset can operate as vision models beyond just video models, we adopt the ImageNet classification task. The last columns of Table~\ref{tab:ac_linprob} and Table~\ref{tab:ac_attnprob} show the results. In general, the models that peform well on appearance-centric action recognition tasks (e.g., K400) also exhibit strong performance on the ImageNet benchmark. However, this tendency changes in some respects. For example, although InternVideo2 significantly outperforms our model on ImageNet, the gap in the video action recognition task significantly decreases (or even reverses). This observation allows us to verify that the proposed method 1) relatively lacks the ability to process static images, and 2) has the capability to utilize motion information for understanding videos in addition to appearance. To advance beyond being just a video model, we need to find a better way to improve the model's ability to understand single images.

\subsection{Temporal Action Localization}
\begin{table}
\centering
\begin{minipage}{0.676\textwidth}
\centering
 \caption{\textbf{Hyperparameters for TAL and TAS evaluation.} }
 \resizebox{\linewidth}{!}{
 \begin{tabular}{l|cc|ccc}
 \hline
 \multirow{2}{*}{Hyperparameter} & \multicolumn{2}{c|}{TAL} & \multicolumn{3}{c}{TAS} \\ \cline{2-6}
 & THUMOS14 & ActivityNet-v1.3 & 50Salads & GTEA & Breakfast \\ \hline
 \multicolumn{3}{l}{\it{\textbf{Backbone Setting}}} \\ \hline
 video FPS & 30 & 30 & 30 & 15 & 15 \\
 embedding type & \multicolumn{2}{c|}{\texttt{Clip Embedding}} & \multicolumn{3}{c}{\texttt{Clip Embedding}} \\
 clip length $T$ (secs) & 0.5 & 0.5 & 1.0 & 0.5 & 0.5 \\ 
 temporal stride $T_s$ (secs) & 0.125 & 0.25 & 0.25 & 0.125 & 0.125 \\ 
 feature post-processing & crop & resize & - & - & - \\
 feature (max) length & 2304 & 192 & \multicolumn{3}{c}{variable} \\ 
 \hline
 \multicolumn{3}{l}{\it{\textbf{Detector Setting}}} \\
 \hline
 batch\_size & 2 & 16 & \multicolumn{3}{c}{1} \\
 epochs & 45 & 10 & \multicolumn{3}{c}{120} \\ 
 warm-up epochs & \multicolumn{2}{c|}{2} & - & - & - \\ 
 lr & 0.001 & 1e-4 & 5e-4 & 5e-4 & 1e-4 \\ 
 lr\_decay & \multicolumn{2}{c|}{cosine annealing} & \multicolumn{3}{c}{reduce on plateau} \\
 weight\_decay & \multicolumn{2}{c|}{0.05} & \multicolumn{3}{c}{1e-4} \\ 
 \hline
 \end{tabular}
}
 \label{tab:tal_tas_hyperparams}
\end{minipage}
\hfill
\hspace{-0.4cm}
\begin{minipage}{0.315\textwidth}
\centering
 \caption{\textbf{Hyperparameters for STAL evaluation.} }
 \label{table:stal_param}
 \resizebox{\linewidth}{!}{
\begin{tabular}{ll}
\hline
  Hyperparameters & AVA v2.2  \\ \hline
\multicolumn{1}{l}{\it{\textbf{Decoder Architecture}}} \\ \hline
\# of layers & 6   \\
hidden size  & 256 \\
MLP dimension   & 2048   \\
box head \# layers & 3   \\
dropout rate & 0.1 \\ 
drop path & 0.1 \\ \hline
\multicolumn{1}{l}{\it{\textbf{Training}}} \\ \hline
optimizer & AdamW  \\
optimizer momemtum & $\beta_1=0.9$, $\beta_2=0.999$ \\
weight\_decay & 1e-5   \\
learning rate   & 1e-4   \\
batch\_size   & 256 \\
warmup epochs   & 10  \\
epochs & 50  \\ \hline
\end{tabular}
}
\end{minipage}
\end{table}
\begingroup
\begin{table}
 \centering
  \caption{\textbf{Temporal action localization evaluation on THUMOS14. } }
\resizebox{0.85\linewidth}{!}{
 \begin{tabular}{l|c|ccc|c|ccc|c}
  \hline
  \multirow{2}{*}{Method} & \multirow{2}{*}{Arch.} & \multicolumn{4}{c|}{Self-contained} & \multicolumn{4}{c}{w/ External Classifier} \\ 
  \cline{3-10}
  & & 0.3 & 0.5 & 0.7 & Avg. & 0.3 & 0.5 & 0.7 & Avg. \\ \hline
UMT$_{s2}$ & ViT-B & 54.60 & 42.18 & 22.17 & 40.34 & 56.39 & 43.23 & 23.27 & 41.66 \\
\rowcolor{tl!50} \adjustbox{valign=c}{\includegraphics[width=0.02\linewidth]{tables/logo.png}} {\ourmodel} & ViT-B & \textbf{71.01} & \textbf{57.06} & \textbf{33.25} & \textbf{54.83} & \textbf{66.09} & \textbf{54.14} & \textbf{33.16} & \textbf{52.05} \\ \hline
UMT$_{s2}$ & ViT-L & 64.17 & 51.79 & 29.61 & 49.43 & 61.53 & 49.22 & 29.32 & 47.31 \\
V-JEPA  & ViT-L & 62.89 & 52.31 & 31.43 & 49.72 & 66.54 & 54.50 & 33.60 & 52.11 \\
\rowcolor{tl!50} \adjustbox{valign=c}{\includegraphics[width=0.02\linewidth]{tables/logo.png}} {\ourmodel} & ViT-L & \textbf{73.61} & \textbf{61.21} & \textbf{38.40} & \textbf{58.75} & \textbf{66.83} & \textbf{55.54} & \textbf{35.88} & \textbf{53.63} \\
  \hline
DFN & ViT-H & 64.54 & 50.79 & 26.59 & 48.17 & 58.11 & 44.44 & 24.40 & 43.21 \\
V-JEPA & ViT-H & 62.98 & 52.48 & 32.19 & 49.91 & 65.31 & 53.79 & 32.94 & 51.54 \\
InternVideo2$_{s2}$ & ViT-g & 81.09 & 69.94 & 44.40 & 66.21 & 68.51 & 58.98 & 37.60 & 55.81 \\
  \hline
 \end{tabular}
}
 \label{tab:tal_thumos14}
\end{table}
\endgroup

\begingroup
\begin{table}
\begin{minipage}{0.67\textwidth}
\centering
 \centering
  \caption{\textbf{Temporal action localization on ActivityNet-v1.3. } \textsuperscript{\textdagger}~indicates results of the models trained including ActivityNet-v1.3. }
\resizebox{1.02\linewidth}{!}{
 \begin{tabular}{l|c|ccc|c|ccc|c}
  \hline
  \multirow{2}{*}{Method} & \multirow{2}{*}{Arch.} & \multicolumn{4}{c|}{Self-contained} & \multicolumn{4}{c}{w/ External Classifier} \\ \cline{3-10}
  & & 0.5 & 0.75 & 0.95 & Avg. & 0.5 & 0.75 & 0.95 & Avg. \\ \hline
UMT$_{s2}$ & ViT-B & 45.82 & 30.32 & 6.23 & 29.86 & 58.12 & 38.23 & 7.55 & 37.73 \\
\rowcolor{tl!50} \adjustbox{valign=c}{\includegraphics[width=0.03\linewidth]{tables/logo.png}} {\ourmodel} & ViT-B & \textbf{49.14} & \textbf{33.06} & \textbf{6.46} & \textbf{32.34} & \textbf{58.60} & \textbf{39.30} & \textbf{8.14} & \textbf{38.56} \\ \hline
UMT$_{s2}$ & ViT-L & 52.06 & 34.62 & 7.24 & 34.07 & 58.98 & 39.40 & 8.09 & 38.70 \\
V-JEPA  & ViT-L & 42.56 & 28.65 & 6.07 & 28.22 & 57.88 & 38.77 & 7.51 & 37.99 \\
\rowcolor{tl!50} \adjustbox{valign=c}{\includegraphics[width=0.03\linewidth]{tables/logo.png}} {\ourmodel} & ViT-L & \textbf{52.96} & \textbf{36.07} & \textbf{7.88} & \textbf{34.98} & \textbf{59.56} & \textbf{40.47} & \textbf{9.06} & \textbf{39.49} \\
  \hline
DFN & ViT-H & 54.22 & 35.79 & 7.25 & 35.33 & 58.62 & 39.34 & 7.61 & 38.36 \\
V-JEPA & ViT-H & 39.39 & 26.76 & 5.60 & 26.18 & 58.49 & 39.33 & 8.23 & 38.46 \\
InternVideo2$_{s2}$ & ViT-g & 58.30\textsuperscript{\textdagger} & 38.85\textsuperscript{\textdagger} & 8.90\textsuperscript{\textdagger} & 38.35\textsuperscript{\textdagger} & 61.36\textsuperscript{\textdagger} & 41.89\textsuperscript{\textdagger} & 9.01\textsuperscript{\textdagger} & 40.82\textsuperscript{\textdagger} \\
  \hline
 \end{tabular}
}
 \label{tab:tal_activitynet}
\end{minipage}
\hspace{0.01\textwidth}
\begin{minipage}{0.307\textwidth}
\centering
 \centering
  \caption{\textbf{Spatio-temporal action localization on AVA v2.2.}}
  \label{table:stal}
\resizebox{\linewidth}{!}{
 \begin{tabular}{l|c|c}
  \hline
  Method & Arch. & fAP@0.5 \\ \hline
UMT$_{s2}$ & ViT-B & \textbf{21.88} \\
\rowcolor{tl!50} \adjustbox{valign=c}{\includegraphics[width=0.06\linewidth]{tables/logo.png}} {\ourmodel} & ViT-B & 20.50 \\ \hline
UMT$_{s2}$ & ViT-L & 26.38 \\
V-JEPA  & ViT-L & 22.57 \\
\rowcolor{tl!50} \adjustbox{valign=c}{\includegraphics[width=0.06\linewidth]{tables/logo.png}} {\ourmodel} & ViT-L & \textbf{27.39} \\
  \hline
DFN & ViT-H & 23.41 \\
V-JEPA & ViT-H & 22.59 \\
InternVideo2$_{s2}$ & ViT-g & 28.68 \\
  \hline
 \end{tabular}
}
 \label{tab:stal}
\end{minipage}
\end{table}
\endgroup

Temporal Action Localization (TAL) is a pivotal task for accurately identifying and understanding human activities within long and intricate videos. This task can be extended to various applications in domains such as autonomous driving, sports analysis, and content-based video retrieval. 
TAL aims to analyze an untrimmed video and precisely identify the temporal intervals and corresponding class labels of distinct actions occurring within the video. We expect that this task can evaluate the models from two perspectives: temporal sensitivity and instance discrimination. Temporal sensitivity requires the model to identify if an action of interest occurs at the current time step. Instance discrimination, on the other hand, requires the ability to distinguish or group frame-wise segments into a complete action instance. Although the task is designed to be motion-centric, we find that the two aspects can be achieved with both appearance and motion capabilities.

\noindent\textbf{Evaluation Settings. } 
For the TAL task, we use the ActivityNet-v1.3~\cite{caba2015activitynet} and THUMOS14~\cite{jiang2014thumos14} datasets, and we adopt ActionFormer~\cite{zhang2022actionformer} as the detection head.
We report the results using two evaluation methods indicated as `Self-contained' and `w/ External Classifier' in Table~\ref{tab:tal_activitynet} and~\ref{tab:tal_thumos14}.
`Self-contained' means that the task requires the model to conduct both classification and regression without any external classifier.
`w/ External Classifier' indicates that the model performs binary classification while an external classifier predicts the action class.
We adopt the external classifier used in \cite{chen2024videomambasuite}.
We evaluate in both ways using various metrics to comprehensively understand VFMs.
We extract temporal features following the \texttt{Multi-Clip Embedding} method described in Section~\ref{subsec:training_details_clip_sampling}.
We divide each video into clips of $T$ seconds with a stride of $T_s$, where $T_s < T$ allows overlap for dense extraction. We then uniformly sample $N$ frames within each clip, with $N$ depending on the encoder configuration. 
Detailed hyperparameters for evaluation are provided in Table~\ref{tab:tal_tas_hyperparams}.
Unless otherwise noted, we follow the configuration of ActionFormer.

\noindent\textbf{Results and Discussion. }
The first point to note is the difference in performance tendencies between ActivityNet-v1.3 and THUMOS14. In ActivityNet-v1.3, the model's performance with an external classifier is higher than that of the self-contained evaluation, whereas in THUMOS14, the opposite trend is observed. This can be attributed to two main factors: the number of classes and the number of training samples. ActivityNet-v1.3 has 200 classes, which is ten times more than THUMOS14 (20 classes). Therefore, requiring a single model to perform both action boundary regression and classification simultaneously demands more representational power than the model's capacity. On the other hand, THUMOS14 has a relatively small training dataset (refer to Table~\ref{tab:eval_benchmark}). As a result, the classification objective provides extra supervision that can compensate for the lack of training samples. V-JEPA uniquely exhibits a different trend from the other models, showing lower self-contained evaluation performance in both THUMOS14 and ActivityNet-v1.3. This can be interpreted as V-JEPA's limited ability to embed appearance information. In fact, the performance drop in ActivityNet-v1.3 is larger than for the other models. Notably, for V-JEPA, even though the performance with the external classifier improves with the ViT-H scale model, the self-contained evaluation performance decreases. This indicates that V-JEPA's motion understanding ability improves, but its appearance understanding ability does not as the model scale increases. This observation aligns with what was noted in Section~\ref{sec:action_recognition}. {\ourmodel} exhibits the best performance across all evaluation methods and benchmarks compared to models of the same scale. Remarkably, {\ourmodel} at the ViT-L scale even outperforms the larger scale models DFN and V-JEPA at the ViT-H scale. This demonstrates that our model possesses robust and generalizable capabilities as a backbone for temporal localization. Moreover, our model shows high performance at strict IoU thresholds, such as 0.95 in ActivityNet, demonstrating its strength in temporal sensitivity. Lastly, InternVideo2 performs best across all scales, especially in self-contained evaluation. This indicates that the TAL task requires a larger model scale than classification alone. Specifically, performing both classification and action boundary regression simultaneously demands more parameters. It is recommended to use a classifier to primarily evaluate motion understanding capability through the TAL task, while for a more comprehensive assessment, including appearance, it is better to use a self-contained approach.

\subsection{Spatio-Temporal Action Localization}
Compared to the previous tasks where the input videos are trimmed or cropped, the spatio-temporal action recognition (STAL) task aims to evaluate the capability of the vision encoder to localize instances and recognize their actions~\cite{yuan2023videoglue}. This task requires the model to not only identify the action taking place but also accurately determine the spatial and temporal locations of the action within an untrimmed video.

\noindent\textbf{Evaluation Settings.} We evaluate the models on the classic STAL dataset AVA v2.2~\cite{gu2018ava}. 
The AVA v2.2 dataset consists of 430 movie clips, each 15 minutes long. 
Instead of annotating all frames, keyframes are annotated every second, resulting in 210,634 labeled frames in the training set and 57,371 in the validation set. 
There are 80 atomic actions labeled for every actor in the clip~\cite{gu2018ava}. 
As a metric, we report the Frame Average Precision (fAP) at an IoU threshold of 0.5 using the latest v2.2 annotations~\cite{gu2018ava}.
In previous works~\cite{zhao2024videoprism,wang2022internvideo}, external ROI headers such as Mask-RCNN~\cite{he2017mask} are used for extracting the bounding box of the instance in the patch-level representation. 
This process suggests that the ability to understand spatial information is limited by whether the patch token has a corresponding spatial feature, not the spatial localizing capacity itself.
Therefore, we decide to adopt the end-to-end STAL framework~\cite{gritsenko2024end}, training the decoder while freezing the vision backbone.
The detailed hyperparameters are shown in Table~\ref{table:stal_param}.

\noindent\textbf{Results and Discussion.} 
We report the evaluation results in Table~\ref{table:stal}. 
In the STAL task, UMT, our model, and InternVideo2 outperform DFN and V-JEPA. 
Even though DFN and V-JEPA show opposite trends in previous evaluations, they exhibit similar performance in this task.
This is because the STAL task consists of two subtasks: localizing the instance and recognizing the action class.
If a model localizes the instance incorrectly, it cannot perform the subsequent task of action recognition effectively. From this perspective, V-JEPA fails to localize the instance correctly, which hinders its ability to solve the recognition task. On the other hand, DFN can localize the instance due to its superior understanding of appearance. However, even if DFN successfully finds the instance, it fails to predict the action class due to an insufficient understanding of temporal information.
Unlike DFN and V-JEPA, models such as UMT, ours, and InternVideo2 are validated to understand both appearance and motion reasonably well, allowing them to solve the problem at a similar level.
From these results, we can verify that 1) the proposed model has the capability to understand spatio-temporal information, and 2) the STAL task in an end-to-end manner can be a proper measure to evaluate the ability to understand video comprehensively.

\subsection{Temporal Action Segmentation}
Temporal Action Segmentation (TAS) is an essential task for the comprehension and analysis of human activities within complex, extended videos, encompassing a diverse array of applications such as video surveillance, video summarization, and skill assessment. 
The TAS task aims to process an untrimmed video input and generate an action sequence of the class label for each frame.

\begingroup
\begin{table}
 \centering
  \caption{\textbf{Temporal action segmentation evaluation on three benchmarks. } }
\resizebox{\linewidth}{!}{
 \begin{tabular}{l|c|ccc|ccc|ccc}
  \hline
  \multirow{2}{*}{Method} & \multirow{2}{*}{Arch.} & \multicolumn{3}{c|}{50Salads} & \multicolumn{3}{c|}{GTEA} & \multicolumn{3}{c}{Breakfast} \\ \cline{3-11}
  & & mF1 & Edit & Acc. & mF1 & Edit & Acc. & mF1 & Edit & Acc. \\ \hline
UMT$_{s2}$ & ViT-B & 70.66 & 70.79 & 77.71 & 80.75 & 83.22 & 75.92 & 31.39 & 43.75 & 32.39 \\
\rowcolor{tl!50} \adjustbox{valign=c}{\includegraphics[width=0.02\linewidth]{tables/logo.png}} {\ourmodel} & ViT-B & \textbf{80.69} & \textbf{76.93} & \textbf{85.83} & \textbf{88.26} & \textbf{87.70} & \textbf{82.94} & \textbf{52.18} & \textbf{62.77} & \textbf{57.86} \\ \hline
UMT$_{s2}$ & ViT-L & 71.20 & 69.23 & 76.60 & 85.91 & 87.62 & 79.69 & 45.25 & 58.11 & 49.04 \\
V-JEPA  & ViT-L & 60.85 & 61.17 & 70.30 & 87.42 & 86.84 & 81.63 & 46.96 & 58.63 & 52.66 \\
\rowcolor{tl!50} \adjustbox{valign=c}{\includegraphics[width=0.02\linewidth]{tables/logo.png}} {\ourmodel} & ViT-L & \textbf{80.60} & \textbf{77.19} & \textbf{84.75} & \textbf{88.43} & \textbf{90.19} & \textbf{82.29} & \textbf{50.66} & \textbf{62.80} & \textbf{54.17} \\
  \hline
DFN & ViT-H & 73.22 & 71.56 & 80.13 & 88.06 & 89.66 & 80.16 & 41.73 & 55.47 & 46.40 \\
V-JEPA & ViT-H & 54.73 & 56.51 & 63.25 & 84.87 & 85.49 & 80.82 & 45.55 & 57.07 & 51.92 \\
InternVideo2$_{s2}$ & ViT-g & 82.00 & 79.04 & 86.74 & 92.96 & 93.08 & 84.76 & 57.38 & 66.98 & 62.38 \\
  \hline
 \end{tabular}
}
  \vspace{10pt}
 \label{tab:tas}
\end{table}
\endgroup
\begin{figure}[t]
    \centering
    \includegraphics[width=\textwidth]{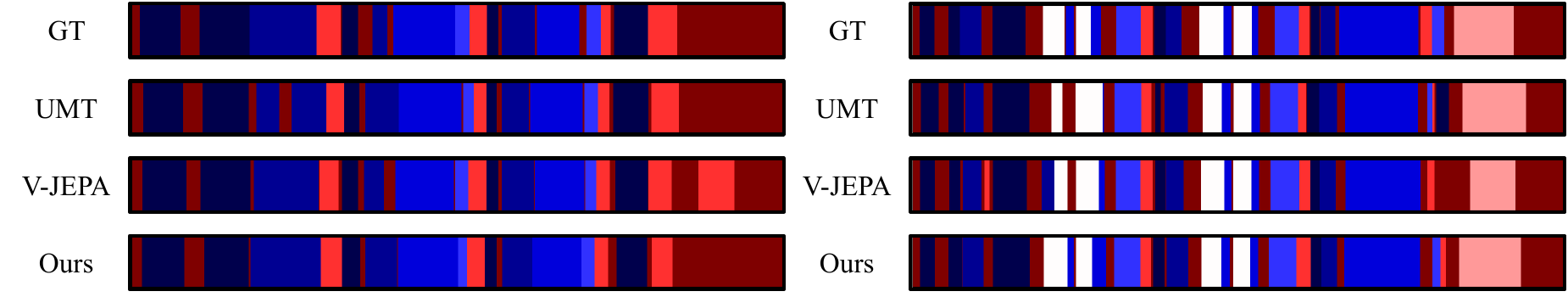}
    \caption{\textbf{Visualization of temporal action segmentation. } The figure shows the qualitative results of two test samples from GTEA. Each action class corresponds to the different color and the x-axis represents time.}
    \label{fig:action_segmentation}
\end{figure}

\noindent\textbf{Evaluation Settings. }
In this paper, we utilize three challenging benchmarks for TAS evaluation: 50Salads~\cite{stein201350salads}, GTEA~\cite{fathi2011gtea}, and Breakfast~\cite{kuehne2014breakfast}.
We employ ASFormer~\cite{yi2021asformer} as the detection head.
The average F1 scores at thresholds of 10, 25, and 50, referred to as `mF1', are reported.
Similar to TAL, we extract features according to the \texttt{Multi-Clip Embedding} method described in Section~\ref{subsec:training_details_clip_sampling} and aggregate them using spatial pooling to retain only the temporal axis.
Further implementation details are provided in Table~\ref{tab:tal_tas_hyperparams}.
Unless otherwise specified, we adhere to the configuration of ASFormer.

\noindent\textbf{Results and Discussion. }
Table~\ref{tab:tas} presents the TAS evaluation results on three benchmarks.
The appearance-centric models, such as UMT and DFN, perform better than the motion-centric model V-JEPA on 50Salads, but they exhibit worse performance on the GTEA and Breakfast datasets.
In contrast, our model surpasses the baselines across all three datasets, demonstrating strong capabilities in both appearance and motion.
Additionally, our model excels in Edit and mF1 scores, indicating that the sequence of predictions closely aligns with the ground-truth instances.
Regarding camera views, 50Salads and GTEA primarily show human hands in top-down and ego-centric perspectives, while Breakfast features a wider view that mostly includes the human body.
In this context, our model outperforms baselines across various views and performs comparably to InternVideo2 in top-down and egocentric views on the 50Salads and GTEA datasets.
Figure~\ref{fig:action_segmentation} illustrates the visualization of two test samples from GTEA for large-scale models: UMT, V-JEPA, and ours.
As depicted, our model produces precise segmentation, while UMT and V-JEPA generate more false positives in the sequence.

\begin{figure*}[t!]
    \centering
    \begin{subfigure}[t]{0.24\textwidth}
        \centering
        \includegraphics[width=\linewidth]{figure/embed_vis/vis_k400_ours.png}
        \caption{{\ourmodel}}
    \end{subfigure}
    \begin{subfigure}[t]{0.24\textwidth}
        \centering
        \includegraphics[width=\linewidth]{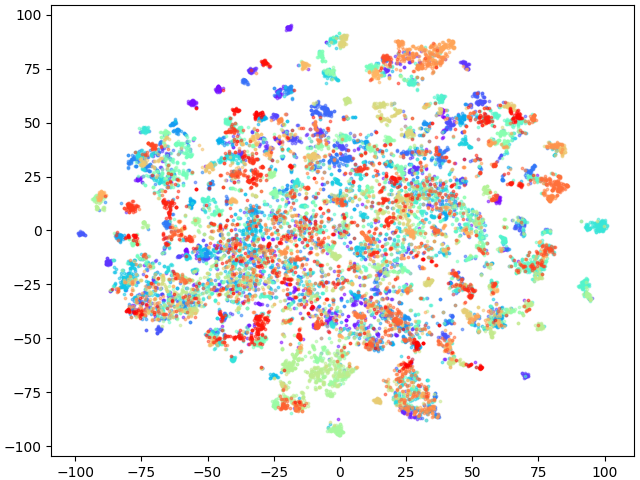}
        \caption{UMT}
    \end{subfigure}
    \begin{subfigure}[t]{0.24\textwidth}
        \centering
        \includegraphics[width=\linewidth]{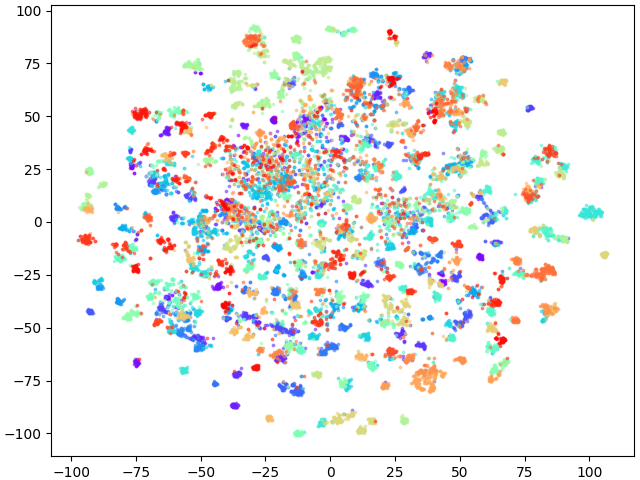}
        \caption{InternVideo2}
    \end{subfigure}
    \begin{subfigure}[t]{0.24\textwidth}
        \centering
        \includegraphics[width=\linewidth]{figure/embed_vis/vis_k400_vjepa.png}
        \caption{V-JEPA}
    \end{subfigure}
    \caption{\textbf{t-SNE visualization of Kinetics-400 validation set.} {\ourmodel}, UMT, and InternVideo2 cluster the samples into several groups while V-JEPA does not provide a clear distinction between groups. We note that all the pretraining dataset of video models includes Kinetics-400 (K400).}
    \label{fig:vis_k400}
\end{figure*}

\begin{figure*}[t!]
    \centering
    \begin{subfigure}[t]{0.24\textwidth}
        \centering
        \includegraphics[width=\linewidth]{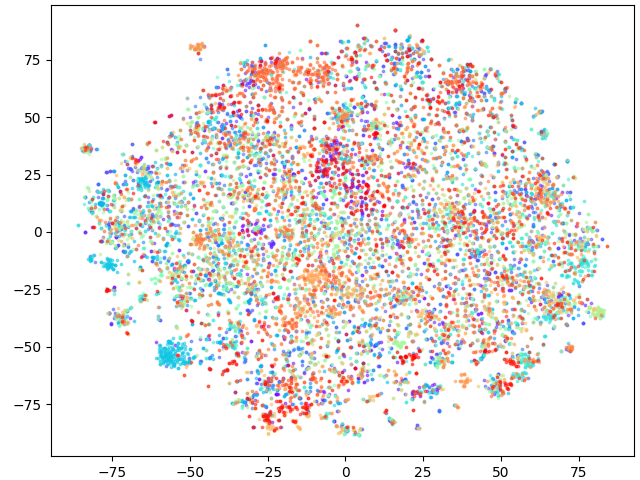}
        \caption{{\ourmodel}}
    \end{subfigure}
    \begin{subfigure}[t]{0.24\textwidth}
        \centering
        \includegraphics[width=\linewidth]{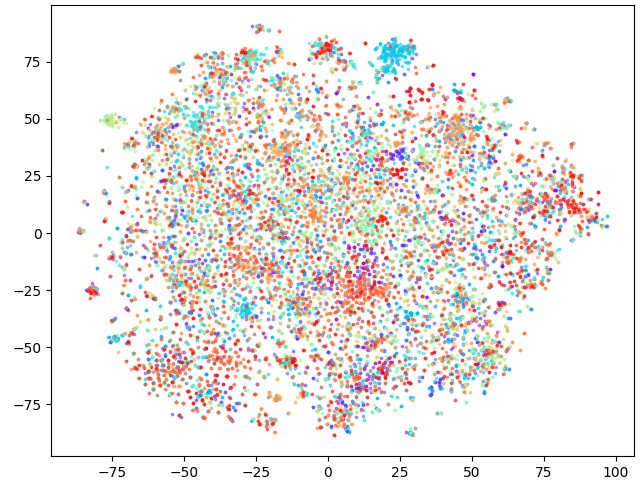}
        \caption{UMT}
    \end{subfigure}
    \begin{subfigure}[t]{0.24\textwidth}
        \centering
        \includegraphics[width=\linewidth]{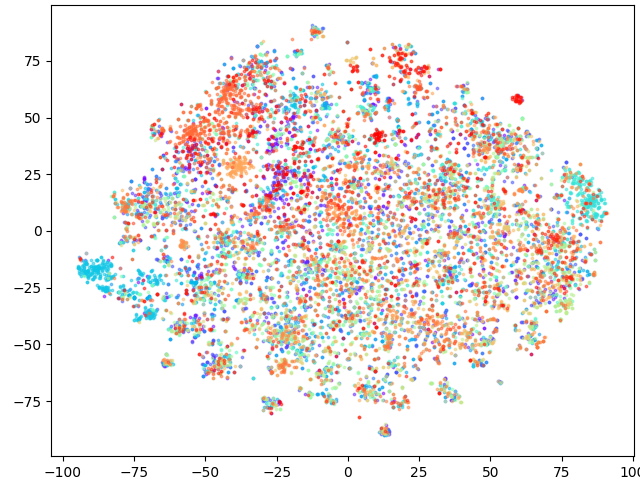}
        \caption{InternVideo2}
    \end{subfigure}
    \begin{subfigure}[t]{0.24\textwidth}
        \centering
        \includegraphics[width=\linewidth]{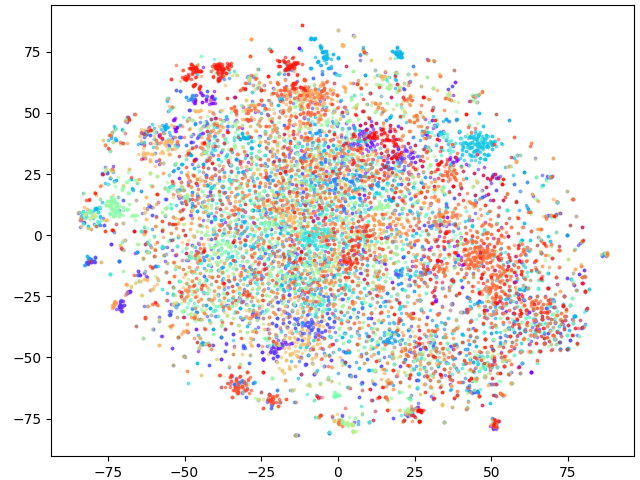}
        \caption{V-JEPA\textsuperscript{\textdagger}}
    \end{subfigure}
    \caption{\textbf{t-SNE visualization of Something-Something-v2 validation set.} 
    It shows uniformly sampled half of the points. All the models fail to provide clear groups because of their low absolute accuracy on SSv2. \textsuperscript{\textdagger} denotes that the pretraining dataset for the model includes the Something-Something-v2 dataset.}
    \label{fig:vis_ssv2}
\end{figure*}

\subsection{Embedding Visualization}
\label{sec:embedding_vis}
\subsubsection{T-SNE Visualization on Benchmark Datasets}
Figures~\ref{fig:vis_k400} and ~\ref{fig:vis_ssv2} visualize the embedding space of the K400 and SSv2 validation sets using t-SNE~\cite{tsne}, with different colors representing different classes. In Figure~\ref{fig:vis_k400}, our model, along with UMT and InternVideo2, demonstrates clusters with clear decision boundaries for most classes. In contrast, V-JEPA fails to cluster the samples into distinct groups, which aligns with the analysis in Section~\ref{sec:action_recognition}. On the other hand, the results in Figure~\ref{fig:vis_ssv2} indicate that none of the models, including V-JEPA, effectively cluster the classes. While Section~\ref{sec:action_recognition} shows that our model and V-JEPA perform relatively better, the absolute performance of classification on the SSv2 dataset is significantly lower compared to the K400 dataset. This discrepancy suggests that none of the models achieve notable clustering results during visualization, highlighting the need for further research to improve performance on motion-centric benchmarks.

\subsubsection{Directional Motion Distinguishability}
\label{sec:forward_reverse}
\begin{figure*}[t!]
    \centering
    \begin{subfigure}[t]{0.24\textwidth}
        \centering
        \includegraphics[width=\linewidth]{figure/updown/vis_up_ours.png}
        \caption{{\ourmodel}}
    \end{subfigure}
    \begin{subfigure}[t]{0.24\textwidth}
        \centering
        \includegraphics[width=\linewidth]{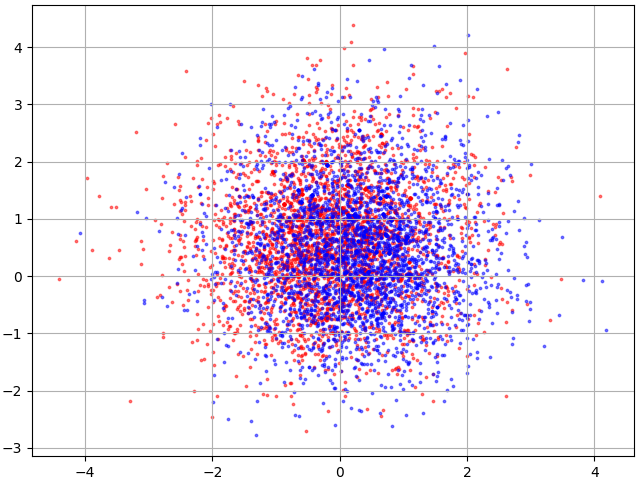}
        \caption{UMT}
    \end{subfigure}
    \begin{subfigure}[t]{0.24\textwidth}
        \centering
        \includegraphics[width=\linewidth]{figure/updown/vis_up_internvideo.png}
        \caption{InternVideo2}
    \end{subfigure}
    \begin{subfigure}[t]{0.24\textwidth}
        \centering
        \includegraphics[width=\linewidth]{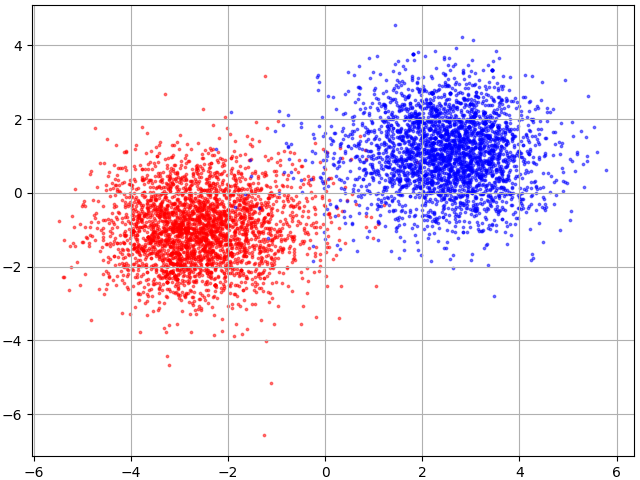}
        \caption{V-JEPA\textsuperscript{\textdagger}}
    \end{subfigure}
    \caption{\textbf{LDA visualization of Something-Something-v2 \textcolor{red}{`Moving something up' class} videos and \textcolor{blue}{their reverse}.} {\ourmodel} and V-JEPA distinguish them well, but the appearance-centric models (UMT and InternVideo2) do not recognize the difference between the forward and reverse video. \textsuperscript{\textdagger} denotes that the pretraining dataset for the model includes the Something-Something-v2 dataset.}
    \label{fig:updown_up}
\end{figure*}

\begin{figure*}[t!]
    \centering
    \begin{subfigure}[t]{0.24\textwidth}
        \centering
        \includegraphics[width=\linewidth]{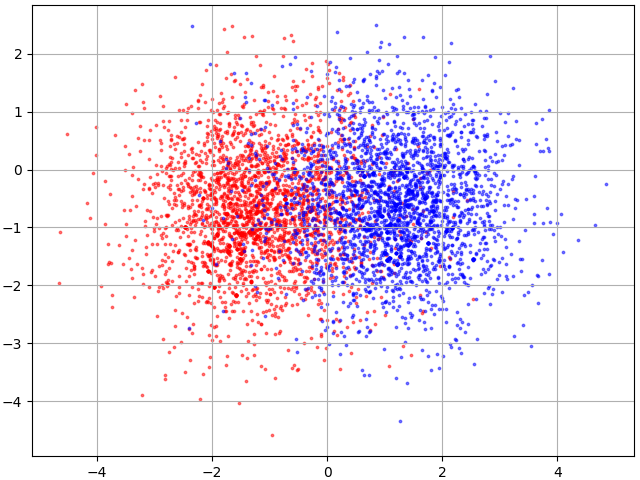}
        \caption{{\ourmodel}}
    \end{subfigure}
    \begin{subfigure}[t]{0.24\textwidth}
        \centering
        \includegraphics[width=\linewidth]{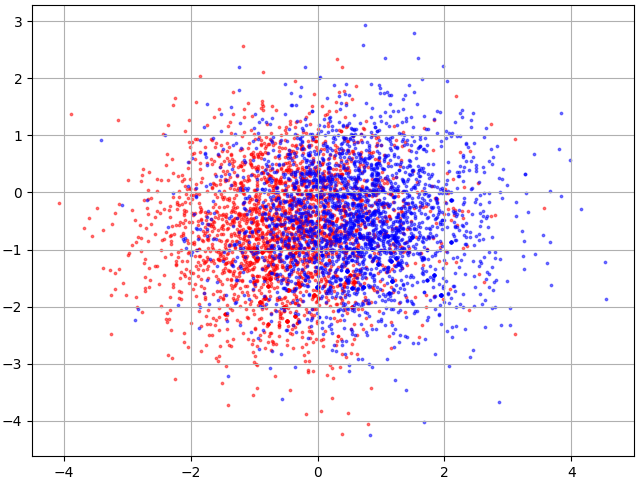}
        \caption{UMT}
    \end{subfigure}
    \begin{subfigure}[t]{0.24\textwidth}
        \centering
        \includegraphics[width=\linewidth]{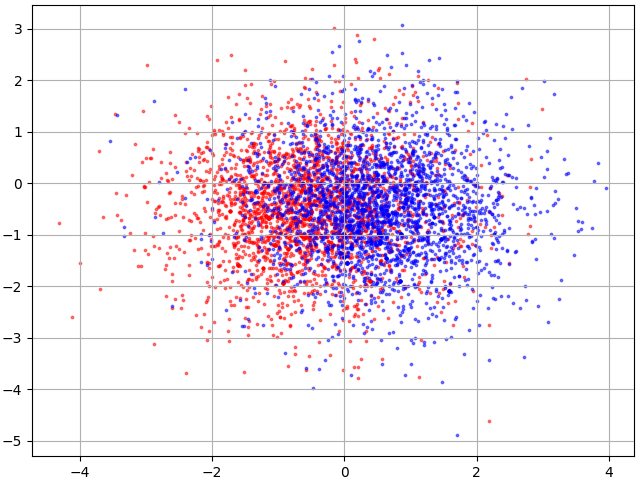}
        \caption{InternVideo2}
    \end{subfigure}
    \begin{subfigure}[t]{0.24\textwidth}
        \centering
        \includegraphics[width=\linewidth]{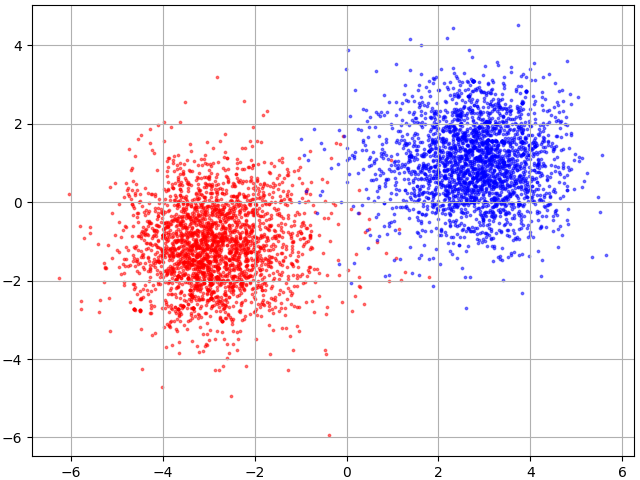}
        \caption{V-JEPA\textsuperscript{\textdagger}}
    \end{subfigure}
    \caption{\textbf{LDA visualization of Something-Something-v2 \textcolor{red}{`Moving something down' class} videos and \textcolor{blue}{their reverse}.} {\ourmodel} and V-JEPA distinguish them well, but the appearance-centric models (UMT and InternVideo2) do not recognize the difference between the forward and reverse video. \textsuperscript{\textdagger} denotes that the pretraining dataset for the model includes the Something-Something-v2 dataset.}
    \label{fig:updown_down}
\end{figure*}

Since the visualization on the motion-centric benchmark dataset (SSv2; Figure~\ref{fig:vis_ssv2}) fails to properly provide information for understanding the models' capabilities, we utilize specific classes within SSv2. Especially, classes like `Moving something up', which require understanding the direction of an object's motion. One way to assess whether a model accurately captures this directional motion is to see if it can distinguish between classes with opposite directions, such as `Moving something up' and `Moving something down'. 
However, there is a limitation in that the appearance of videos in each class differs. To control the difference in appearance, we visualize the embeddings of the `Moving something up' class and its reversed (temporally flipped) version. We then use Linear Discriminant Analysis~(LDA) to visualize whether these two embeddings can be distinguished. This allows us to determine whether a model can encode the direction of motion in the embedding vector without relying only on appearance. Figure~\ref{fig:updown_up} presents the analysis for the `Moving something up' class, and Figure~\ref{fig:updown_down} shows the analysis for the `Moving something down' class. The results indicate that V-JEPA, trained on the SSv2 dataset, shows the highest distinguishability between forward and reverse videos. Our model also demonstrates a good separation between the two. In contrast, UMT and InternVideo2 provide non-separable embeddings, suggesting that while these models have strong appearance understanding capabilities, they fail to adequately distinguish between the forward and reverse versions of the same video. This conclusion indicates that UMT and InternVideo2 do not effectively capture the directional motion information.

\section{Future Directions}

\noindent\textbf{Scaling Up.}
In this technical report, we focus on two scales of models: ViT-B and ViT-L. Our model, at the size of ViT-L, already demonstrates performance comparable to larger scales such as ViT-H or ViT-g. Therefore, further scaling of the model has the potential to enhance performance even more. Along with increasing model size, another future direction involves using large-scale datasets collected in-house to develop more general and robust models.

\noindent\textbf{Image Embedding.}
As mentioned in Section~\ref{sec:imagenet}, our model currently exhibits limited image embedding capability. Since an image can be considered a single-frame video, and video understanding is an extension of image understanding, improving our model's image embedding capacity is crucial. Enhancing this aspect will enable the model to transition from a video foundation model to a more versatile visual understanding model, thereby broadening its potential applications.

\noindent\textbf{Expanding Modality}
In addition to the tasks covered in this technical report, there are various other video-related tasks, particularly, multimodal tasks such as retrieval and captioning. To align the model with the text domain while maintaining all its unimodal capabilities, further in-depth research is necessary. Additionally, improvements are needed to develop a more effective vision encoder for video-language models. Proper evaluation methodologies for these tasks must also be proposed to ensure comprehensive assessment and validation.

\section{Conclusion}
In this technical report, we highlight appearance and motion as the two most critical elements of video understanding. Accordingly, we present methodologies for measuring each capability using various tasks and demonstrate through experiments that existing models have limitations in generally satisfying both capabilities. We then propose {\ourmodel}, a video foundation model that is robust in both motion and appearance. We expect our model and the embeddings obtained through it to be actively utilized and researched in various downstream tasks. Furthermore, we hope that the evaluation and analysis methods proposed in this technical report will be actively used in the video foundation model domain and will serve as a guiding direction for the video understanding field.

\newpage
\section*{Authorship}
This work was achieved through the combined efforts~(equal contribution) of the core contributors, with significant support from the Twelve Labs ML Research and ML Data teams.

\subsection*{Core Contributors}
Hyeongmin Lee, \textit{Research Scientist} \\
Jin-Young Kim, \textit{Research Scientist} \\
Kyungjune Baek, \textit{Research Scientist} \\
Jihwan Kim, \textit{Research Scientist} \\
Aiden Lee, \textit{CTO}

\subsection*{Contributors \footnote{\label{footnote:alphabet}The author list sorted alphabetically.}}

William (Hyojun) Go, \textit{Research Scientist} \\
Mars (Seongsu) Ha, \textit{Research Scientist} \\
Cooper (Seokjin) Han, \textit{Research Scientist} \\
Flynn (Jiho) Jang, \textit{Research Scientist} \\
Ray (Raehyuk) Jung, \textit{Research Scientist} \\
Leo (Daewoo) Kim, \textit{Research Scientist} \\
Daniel (GeunOh) Kim, \textit{ML Data Engineer} \\
Max (JongMok) Kim, \textit{Research Scientist} \\
Jeff (Jongseok) Kim, \textit{Research Scientist} \\
Jayden (Junwan) Kim, \textit{Research Intern} \\
Ian (Soonwoo) Kwon, \textit{Research Scientist} \\
Aaron (Jangwon) Lee, \textit{ML Data Intern} \\
Kyle (Seungjoon) Park, \textit{Research Scientist} \\
Calvin (Minjoon) Seo,  \textit{Chief Scientist} \\
Jay Suh, \textit{ML Data Engineer} \\
Jay (Jaehyuk) Yi, \textit{Research Scientist} \\

\subsection*{Acknowledgement}
We thank Jae Lee (CEO) and the leadership team for the support in our research, and the product and go-to-market team at Twelve Labs for their inspiration and guidance. Special thanks to the engineering team for their essential support, and to Sangdoo Yun from Naver AI Lab for his valuable feedback.

\newpage
\bibliography{main}
\bibliographystyle{plain}

\newpage


\end{document}